\documentclass[10pt,conference]{IEEEtran}
\IEEEoverridecommandlockouts
\usepackage{cite}
\usepackage{amsmath,amssymb,amsfonts}
\usepackage{graphicx}
\usepackage{textcomp}
\usepackage{xcolor}
\usepackage{hyperref}

\usepackage{listings}
\usepackage{courier} 
\usepackage{times}
\usepackage{latexsym}
\usepackage{stfloats} 
\usepackage{adjustbox} 
\usepackage{tcolorbox}

\usepackage{subcaption} 

\usepackage{algorithm}
\usepackage{algpseudocode}
\usepackage{amsmath}

\usepackage{booktabs}
\usepackage{amssymb} 
\usepackage{booktabs}  
\usepackage{pifont}    

\usepackage[T1]{fontenc}

\usepackage[utf8]{inputenc}

\usepackage{microtype}

\usepackage{inconsolata}

\usepackage{tcolorbox}

\usepackage{graphicx}
\usepackage{amsmath}
\usepackage{amssymb}
\usepackage{booktabs}
\usepackage{tabularx}
\usepackage{array}
\usepackage{subcaption}
\usepackage[font=small,labelfont=bf,width=\textwidth]{caption}
\usepackage{algorithm}

\def\BibTeX{{\rm B\kern-.05em{\sc i\kern-.025em b}\kern-.08em
    T\kern-.1667em\lower.7ex\hbox{E}\kern-.125emX}}
\begin{document}

\title{Flying Pigs, \textit{FaR} and Beyond: Evaluating LLM Reasoning in Counterfactual Worlds\\
}

\author{\IEEEauthorblockN{\makebox[\dimexpr 0.33\textwidth\relax][c]{Anish R Joishy*}}
\IEEEauthorblockA{\textit{IIIT Hyderabad} \\
anish.joishy@research.iiit.ac.in}
\and
\IEEEauthorblockN{\makebox[\dimexpr 0.33\textwidth\relax][c]{Ishwar B Balappanawar*}}
\IEEEauthorblockA{\textit{IIIT Hyderabad} \\
ishwar.balappanawar@students.iiit.ac.in}
\and
\IEEEauthorblockN{\makebox[\dimexpr 0.33\textwidth\relax][c]{Vamshi Krishna Bonagiri*}}
\IEEEauthorblockA{\textit{IIIT Hyderabad} \\
vamshi.b@research.iiit.ac.in}
\and
\IEEEauthorblockN{\makebox[\dimexpr 0.33\textwidth\relax][c]{Manas Gaur}}
\IEEEauthorblockA{\textit{University of Maryland,} \\
\textit{Baltimore County} \\
manas@umbc.edu}
\and
\IEEEauthorblockN{\makebox[\dimexpr 0.33\textwidth\relax][c]{Krishnaprasad Thirunarayan}}
\IEEEauthorblockA{\textit{Wright State University} \\
t.k.prasad@wright.edu}
\and
\IEEEauthorblockN{\makebox[\dimexpr 0.33\textwidth\relax][c]{Ponnurangam Kumaraguru}}
\IEEEauthorblockA{\textit{IIIT Hyderabad} \\
pk.guru@iiit.ac.in}
}


\maketitle

\begin{abstract}
A fundamental challenge in reasoning is navigating hypothetical, counterfactual worlds where logic may conflict with ingrained knowledge. We investigate this frontier for Large Language Models (LLMs) by asking: Can LLMs reason logically when the context contradicts their parametric knowledge? To facilitate a systematic analysis, we first introduce \textit{CounterLogic}, a benchmark specifically designed to disentangle logical validity from knowledge alignment. Evaluation of 11 LLMs across six diverse reasoning datasets reveals a consistent failure: model accuracy plummets by an average of 14\% in counterfactual scenarios compared to knowledge-aligned ones. We hypothesize that this gap stems not from a flaw in logical processing, but from an inability to manage the cognitive conflict between context and knowledge. Inspired by human metacognition, we propose a simple yet powerful intervention: \textit{Flag \& Reason (FaR)}, where models are first prompted to \textit{flag} potential knowledge conflicts \textit{before} they reason. This metacognitive step is highly effective, narrowing the performance gap to just 7\% and increasing overall accuracy by 4\%. Our findings diagnose and study a critical limitation in modern LLMs’ reasoning and demonstrate how metacognitive awareness can make them more robust and reliable thinkers\footnote{Our data and code are available at \href{https://github.com/Sukin272/CounterLogic.git}{https://github.com/Sukin272/CounterLogic.git} \\ 
* Authors contributed equally to this work}.
\end{abstract}


\section{Introduction}

\begin{figure}[t]
\centering
\captionsetup{width=\linewidth}\includegraphics[width=\columnwidth]{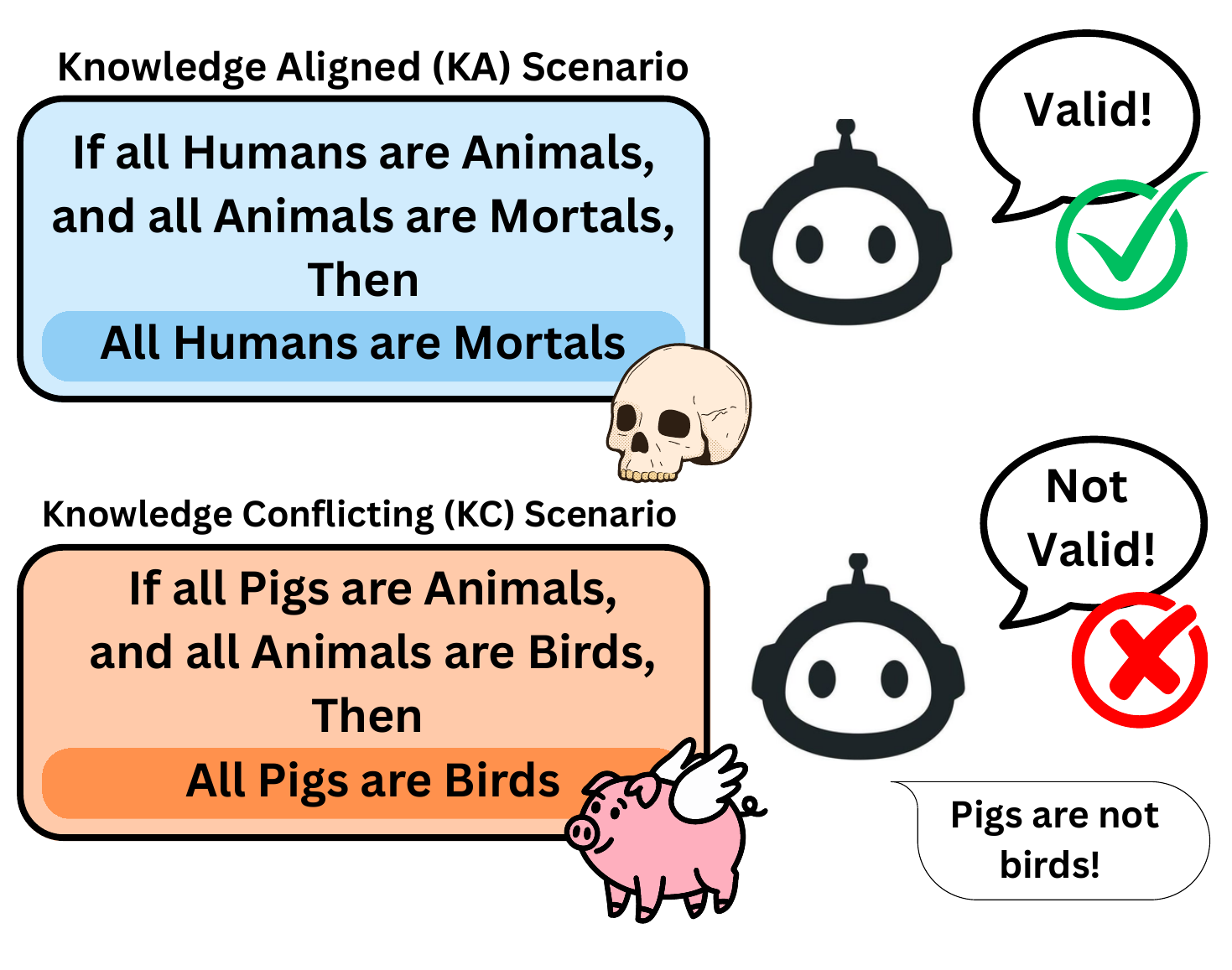}
\caption{When asked to reason solely based on the given premises, LLMs reliably validate arguments that are \textbf{knowledge-aligned} (top), but their performance degrades on logically identical arguments that are \textbf{knowledge-conflicting} (bottom). This is because their parametric knowledge interferes with the reasoning process (e.g., ``Pigs are not birds'').}
\label{fig:motivation}
\end{figure}

LLMs have demonstrated remarkable reasoning capabilities across diverse domains, exhibiting proficiency in tasks ranging from elementary problem solving to complex multi-step reasoning challenges \cite{brown2020language,kojima2022large,schick2021small,wei2023chainofthoughtpromptingelicitsreasoning, zhou2023least,patel2024multi}. Despite these advances, they often exhibit a significant performance degradation (see Fig~\ref{fig:motivation}) when reasoning with information that conflicts with their parametric knowledge  (knowledge conflicts) obtained during training \cite{dasgupta_language_2024,jin2024tug,lampinen2024language,  su2024conflictbank,    wang_resolving_2024,wu_reasoning_2024,xu_knowledge_2024}.

The ability to reason effectively in scenarios with potentially conflicting information is crucial for deploying LLMs in real-world applications where they must process information that may be novel, unexpected, or even contradictory to their training data \cite{wang_resolving_2024}. For instance, engaging in scientific exploration or understanding historical debates requires reasoning from a counterfactual premise, such as: ``Assuming the sun revolves around the Earth, describe the predicted path of Mars in the night sky.'' Such situations are central to creative and scientific thought \cite{pearl2018book}, and failure to reason with them could lead to unreliable performance \cite{zhang2024bridging}. Additionally, prior research also suggests that evaluating reasoning in counterfactual situations may serve as a more robust assessment of a model's reasoning capabilities \cite{wu_reasoning_2024}, as standard reasoning tasks can potentially be \textit{hacked} through pattern matching \cite{lewis-2024-counterfactual,liu_untangle_2024, kaushik-etal-2020-learning,mccoy-etal-2019-right,wu_reasoning_2024}.

While knowledge conflicts are actively studied, prior investigations have focused on relatively simple tasks involving information extraction or single-step reasoning (example: Who is the current president of the USA?) \cite{xie_adaptive_2024}. Consequently, existing benchmarks either test complex logical reasoning without controlling for knowledge conflicts, or test simple knowledge conflicts with shallow, single-step reasoning. As summarized in Table \ref{tab:dataset-comparison}, no benchmark has been designed to systematically probe multi-step logical reasoning under the stress of counterfactuals.

To address this gap, we introduce \textit{CounterLogic}, a benchmark designed to disentangle logical validity from knowledge alignment. Our framework is easily extendable and currently implements 9 logical schemas with problems that systematically scale in complexity from 1 to 9 reasoning steps. Using \textit{CounterLogic} and five other datasets (see Table \ref{tab:dataset-comparison}), our large-scale evaluation of 11 LLMs reveals a pervasive failure: model accuracy plummets by an average of 14\% in counterfactual scenarios. Further analysis reveals that this performance gap is not exacerbated by the number of reasoning steps, but is instead critically dependent on the logical schema. Specifically, the gap soars past 30\% on tasks requiring negation, while being minimal on simpler forms.


We hypothesize that this failure stems not just from an inability to reason, but from an inability to manage the cognitive conflict that arises when premises contradict parametric knowledge. Inspired by human metacognition, the ability to reflect on one's own thinking to resolve such conflicts \cite{fletcher2012metacognition}, we propose a simple yet powerful intervention called  \textit{Flag \& Reason (FaR)}. Before reasoning, we prompt the model to first \textit{flag} whether a conclusion is factually believable, forcing it to confront the knowledge conflict. Our experiments show \textit{FaR} is highly effective, narrowing the performance gap to just 7\% and increasing overall accuracy by 4\% on average. Our analysis of the models' chain-of-thought reveals that this intervention works by anchoring the reasoning process in the initial acknowledgment of the conflict, preventing knowledge bias from derailing the subsequent logical steps.


In this paper, we first use \textit{CounterLogic} (Section \ref{sec:counterlogic}) and other benchmarks to diagnose a pervasive failure in LLM reasoning (Section \ref{sec:findings}). We then propose a metacognitive intervention, \textit{FaR}, to address this failure (Section \ref{sec:method_and_results}) and analyse its effectiveness.


Our contributions can be summarized as follows:
\begin{enumerate}
    \item We introduce \textit{CounterLogic}, a balanced benchmark for evaluating multi-step logical reasoning in knowledge-conflicting scenarios.
    \item We quantify and analyse a pervasive reasoning failure in LLMs: a significant and systematic performance drop when logical validity clashes with a model's parametric knowledge.
    \item We propose and discuss \textit{Flag \& Reason (FaR)}, a simple, metacognitive intervention that significantly mitigates this failure, narrowing the identified performance gap and increasing overall accuracy. 
\end{enumerate}

\begin{table}[t]
  \centering
  \small 
  \setlength{\tabcolsep}{2pt} 
  \begin{tabularx}{\linewidth}{@{}l *{4}{>{\centering\arraybackslash}X}@{}}
    \toprule
    \textbf{Dataset} & \textbf{Size} & \textbf{\# Steps} & \textbf{Conflict} & \textbf{Balance} \\
    \midrule
    LogicBench\\ \cite{parmar2024logicbench} & 2,020 & 1 $\sim$ 5 & \texttimes & \texttimes 
    \\
    FOLIO \\\cite{han2024folionaturallanguagereasoning} & 1,435 & 0 $\sim$ 7 & \texttimes & \texttimes \\
    KNOT \\ \cite{liu_untangle_2024} & 5,500 & 1 $\sim$ 2 & \checkmark & \texttimes \\
    Syllogistic \\ \cite{bertolazzi2024systematicanalysislargelanguage} & 2,120 & 2 & \checkmark & \texttimes \\
    \midrule
    \textbf{CounterLogic (Ours)} & 1,800\textsuperscript{*} & 2 $\sim$ 9\textsuperscript{*} & \checkmark & \checkmark \\
    \bottomrule
  \end{tabularx}
  \captionsetup{width=\linewidth}
  \caption{Comparison of logical reasoning benchmarks. \textit{CounterLogic} uniquely combines multi-step reasoning with knowledge-conflicting scenarios while maintaining balance in labels. This enables rigorous evaluation of how parametric knowledge affects LLMs' logical reasoning capabilities, addressing limitations in existing benchmarks that typically lack proper balance across important evaluation dimensions. \textit{\textsuperscript{*}We provide a dataset generation framework and the numbers can be easily changed as per the requirements. }
  }
  \label{tab:dataset-comparison}
  \vspace{-5pt}
\end{table}

\section{Related Work}


\subsection{Logical Reasoning in LLMs}
Recent advancements have significantly improved the reasoning capabilities of LLMs~. Techniques such as chain-of-thought prompting guide models to articulate intermediate reasoning steps \cite{wei2023chainofthoughtpromptingelicitsreasoning}, zero-shot reasoning allows them to solve problems without task-specific examples \cite{kojima2022large}, and tree-of-thought methods enable the exploration of multiple reasoning paths \cite{yao2023tree}. While these approaches have improved performance on various benchmarks \cite{clark2020transformers, parmar2024logicbench}, studies reveal that LLMs still exhibit systematic errors that often mirror human cognitive biases \cite{dasgupta_language_2024, eisape_systematic_2024}.

Specifically, research into the limitations of logical reasoning shows that models struggle with operations involving negation, quantifiers, and abstract variables \cite{ bertolazzi2024systematicanalysislargelanguage,dasgupta_language_2024, singhprobing}. Performance particularly degrades when reasoning with counterfactual information that contradicts their training data \cite{chen2025counterbench, wu_reasoning_2024}, and they often handle logically equivalent problems inconsistently when presented in different formats \cite{estermann2025reasoning}. Current approaches to address these issues include integrating symbolic representations into the reasoning chain \cite{xu2024faithful} and using verification mechanisms to validate reasoning against formal rules \cite{toroghi2024verifiable,vacareanu2024general}.

\subsection{Knowledge Conflicts and Counterfactual Reasoning}
LLMs store vast amounts of factual knowledge within their parameters, which can create challenges when they encounter conflicting information in a given context \cite{petroni_language_2019, roberts_how_2020}. These knowledge conflicts have been categorized based on their origin, such as between the context and the model's parametric memory \cite{xu_knowledge_2024}. Larger models often default to their parametric knowledge over conflicting contextual evidence, though this can vary based on the coherence and perceived reliability of the source \cite{longpre_entity-based_2021, Wangetal2023}.

This tension is especially pronounced in counterfactual reasoning, where models must follow premises that explicitly contradict established facts \cite{chen2025counterbench,wei2023chainofthoughtpromptingelicitsreasoning}. The performance drop in these scenarios is primarily attributed to the conflict between the model's stored knowledge and the counterfactual assertions it is asked to accept \cite{lin2025generative}. Existing mitigation strategies often involve data augmentation with counterfactual examples or specialized prompting techniques \cite{chen2023discodistillingcounterfactualslarge,neeman2022disentqa, xie_adaptive_2024}. However, these methods have not fully resolved the core issue, particularly in tasks requiring multi-step logical deduction \cite{patel2024multi}.

\subsection{Metacognition and Human Reasoning}
Human reasoning is subject to well-documented cognitive biases, such as the ``belief bias,'' where the believability of a conclusion influences the judgment of an argument's logical validity \cite{markovits_belief-bias_1989}. This bias can create an ``illusion of objectivity,'' where individuals are confident in their biased reasoning \cite{kunda1990case, trippas2014fluency}. LLMs exhibit similar patterns, performing better when a statement's content aligns with the knowledge they gain when training i.e. their parametric knowledge and struggling with implausible scenarios, thus mirroring the systematic errors seen in humans \cite{dasgupta_language_2024, eisape_systematic_2024, macmillan-scott2024irrationality}.

In humans, metacognitive strategies can improve logical reasoning by creating a distinction between evaluating a belief and making a logical inference. In essence, separating ``what I know'' from ``what follows logically'' \cite{douven2018conditionals}. Similar metacognitive-like capabilities are an emerging area of research for LLMs. These include prompting models to estimate their own uncertainty \cite{zhou_navigating_2023}, perform self-evaluation to critique their own reasoning \cite{wang_resolving_2024}, or explicitly identify when a premise conflicts with their knowledge \cite{chen_say_2023}. This parallel suggests a promising direction for enhancing the robustness of logical reasoning in language models, which is the approach our work investigates.

\begin{figure*}[t]
  \centering
  \includegraphics[width=\textwidth]{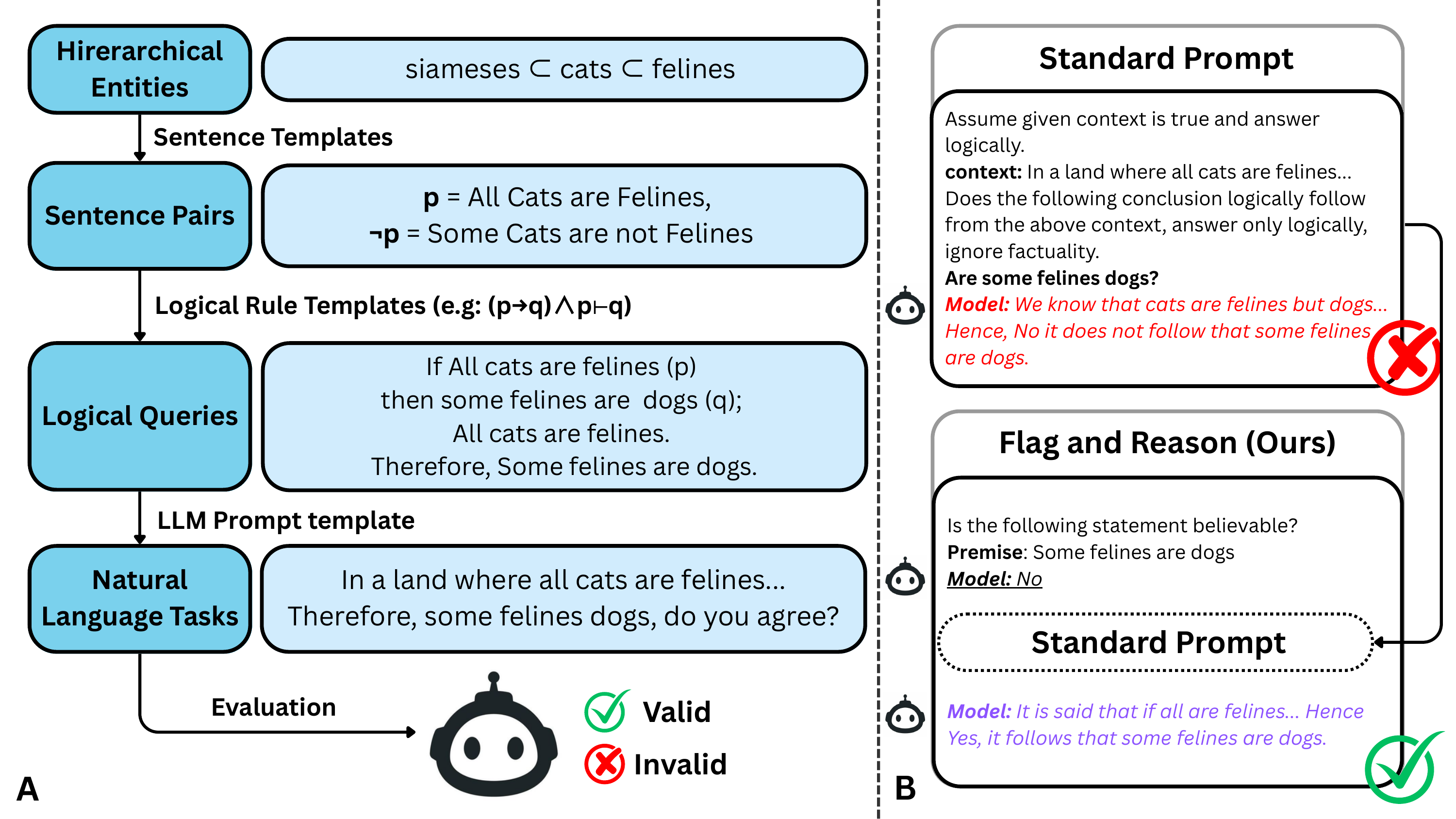}
  \caption{(A) \textbf{Dataset Preparation}:  The dataset features hierarchical entity triples (e.g., siameses $\subset$ cats $\subset$ felines) mapped to 8 logical sentence templates across 9 logical schemas (see Appendix \ref{app:counterlogic}). Each example is balanced across validity (50\% valid/invalid) and believability (50\% aligned/conflicting), with ground truth annotations for both dimensions. The dataset construction combines subset relationships with propositional logic forms such as Modus Ponens, Hypothetical Syllogism, etc. to systematically evaluate knowledge-logic interactions. (B) \textbf{FaR method}: While the standard prompt simply presents LLMs with a counterfactual context followed by related questions, our \textit{FaR} approach first engages the model metacognitively by eliciting its responses to knowledge-alignment questions. This could be as simple as asking whether a given statement is true.}
  \label{fig:crownjewel}
\end{figure*}

\section{The CounterLogic Dataset}
\label{sec:counterlogic}

Existing benchmarks in evaluating LLM reasoning \cite{kojima2022large,wei2023chainofthoughtpromptingelicitsreasoning} fail to systematically disentangle logical validity from knowledge alignment. As highlighted in Table~\ref{tab:dataset-comparison}, current datasets either focus on complex logical structures while ignoring knowledge conflicts like LogicBench \cite{parmar2024logicbench} and FOLIO \cite{han2024folionaturallanguagereasoning}, or test simple knowledge conflicts single-step reasoning like KNOT \cite{liu_untangle_2024} and Reasoning \& Reciting \cite{wu_reasoning_2024}.

To fill this critical gap, we introduce \textit{CounterLogic}, a benchmark built upon a synthetic generation framework designed to enable a controlled evaluation of how parametric knowledge interferes with mulit-step reasoning. The benchmark is easily extendable, contains 1,800 examples across 9 logical schemas shown in Table~\ref{tab:logic-forms}, and features problems that scale in complexity from 1 to 9 reasoning steps.

\begin{figure*}[htbp]
  \centering
  \includegraphics[width=1\linewidth]{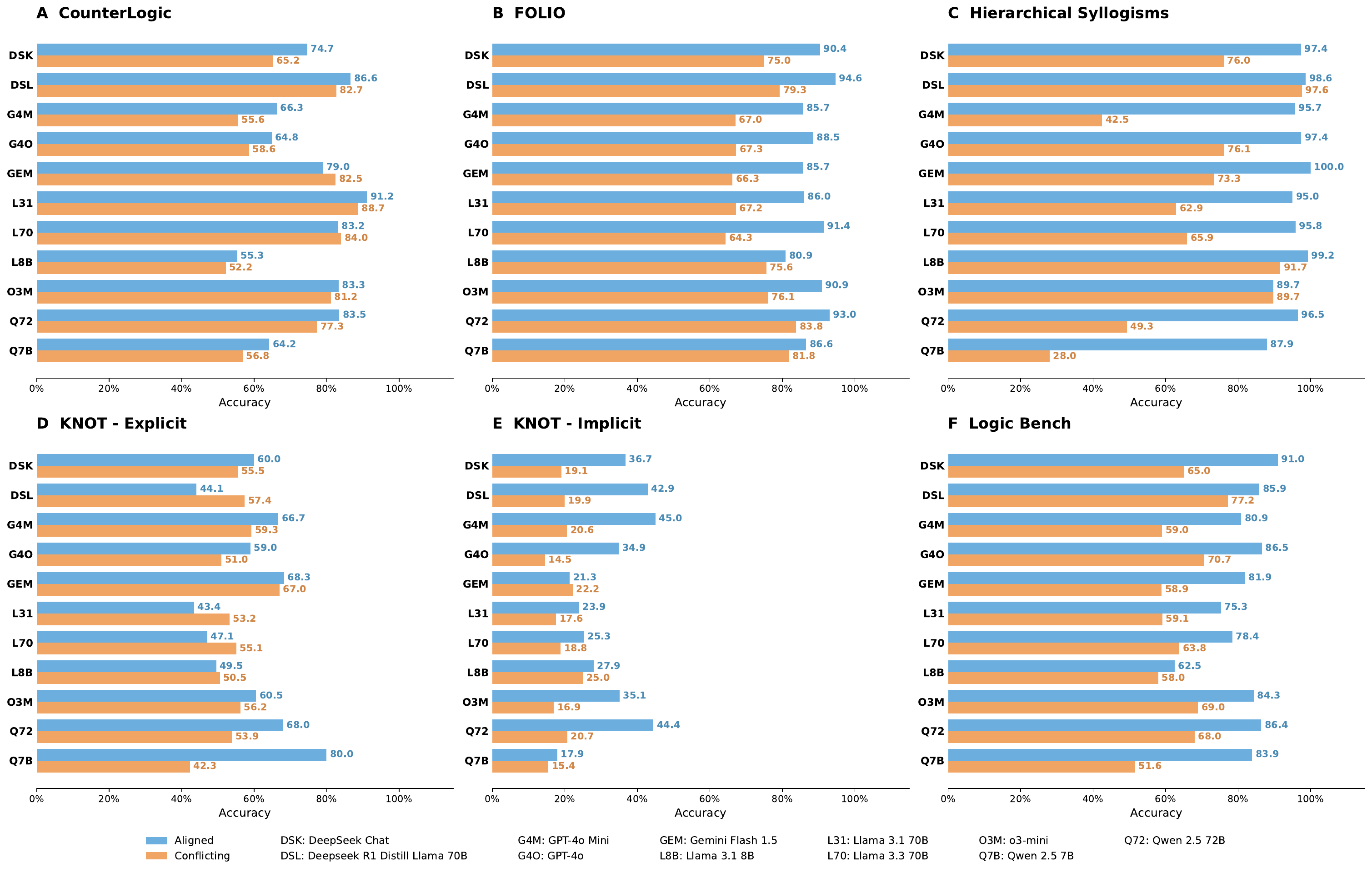}
  
  \captionsetup{width=\linewidth}
  \caption{%
    \textbf{Average accuracy comparison for the Aligned and Conflicting Scenarios.}
    This plot shows comparison between Knowledge-Aligned and Knowledge-Conflicting scenarios where the ground truth is valid. Blue bars represent Knowledge-Aligned example accuracy, while orange bars indicate Knowledge-Conflicting ones. We can see a clear performance gap between these scenarios across all the models and datasets with the average accuracy difference of 14\% across all models and datasets with Hierarchical Syllogisms (C) showing the maximum difference of 27\% (Table \ref{tab:model_performance_valid_base}). y-axis represents the following models - DSK: DeepSeek Chat, DSL: Deepseek R1 Distill Llama 70B, G4M: GPT-4o Mini, G4O: GPT-4o, GEM: Google Gemini Flash 1.5, L8B: Llama 3.1 8B Instruct, L31: Llama 3.1 70B Instruct, L70: Llama 3.3 70B Instruct, O3M: o3-mini, Q7B: Qwen 2.5 7B Instruct, and Q72: Qwen 2.5 72B Instruct.
    Each of the above graphs represents the performance on the following datasets - A: CounterLogic, B: FOLIO, C: Hierarchical Syllogisms, D: KNOT-Explicit, E: KNOT-Implicit, F: Logic Bench.
  }
  \label{fig:combined-gap-analysis}
\end{figure*}

\subsection{Dataset Construction} 

\textbf{Hierarchical Entity Perturbation:} We begin with a set of hierarchical entity triples $(a, b, c)$ representing strict subset relationships ($a \subset b \subset c$), such as \textit{siameses} $\subset$ \textit{cats} $\subset$ \textit{felines}.

\textbf{Sentence Generation:} Entities are inserted into four templates that form complementary logical pairs ($S$ and $\neg S$), such as ``All $\{A\}$ are $\{B\}$'' and ``Some $\{A\}$ are not $\{B\}$''. This process yields atomic propositions with controlled factual accuracy. To ensure diversity, we balance the entity relationships: 50\% use a correct hierarchy (e.g., \textit{siameses} $\subset$ \textit{cats}) and 50\% use an inverted one (e.g., \textit{cats} $\subset$ \textit{siameses}). More details about the sentence generation are in Appendix~\ref{app:entity-triples}.

\textbf{Logical Schema Integration:} Inspired by LogicBench \cite{parmar2024logicbench}, the sentence pairs generated are then integrated as propositions into 9 formal logical schemas: Modus Ponens (MP), Modus Tollens (MT), Hypothetical Syllogism (HS), Disjunctive Syllogism (DS), Constructive Dilemma (CD), Destructive Dilemma (DD), Bidirectional Dilemma (BD), Commutation (CT), and Material Implication (MI) (see Table~\ref{tab:logic-forms} for formal definitions).

\textbf{Balanced Label Assignment:} To enable a rigorous and unbiased evaluation, we ensure the dataset is balanced by creating binary question-answer tasks that systematically control two key dimensions: \textbf{(1) Logical Validity:} We generate logically \textit{Valid} instances where the conclusion logically follows from the premises, and \textit{Invalid} instances where the conclusion is a non-sequitur, violating the logical form.
\textbf{(2) Knowledge Alignment:} Similar to the examples in Fig~\ref{fig:motivation}, we generate \textit{Knowledge-Aligned} instances where the conclusion is factually correct (aligned with parametric knowledge) based on the ground-truth entity hierarchies, and \textit{Knowledge-Conflicting} instances where the conclusion is factually incorrect (contradicting parametric knowledge).

Examples for the above are given in the Appendix \ref{app:dataset stats}. To prevent models from simply memorizing logical forms \cite{xie2024memorization, wu_reasoning_2024}, we use GPT-4o to paraphrase the final logical queries into natural language questions (Appendix~\ref{app:NL-conversion}).

\textbf{Varying Reasoning Depth:} For schemas involving implication ($A \to B$), we create deeper reasoning chains by introducing intermediate steps: $A \to X_1 \to \dots \to X_i \to B$. This allows for a fine-grained analysis of how reasoning depth interacts with the challenges posed by knowledge conflicts, a feature absent in prior benchmarks (see Appendix \ref{app:varying-depth}). 
\label{reasoning-depth-var}

\begin{figure}
    \centering
    \includegraphics[width=1\linewidth]{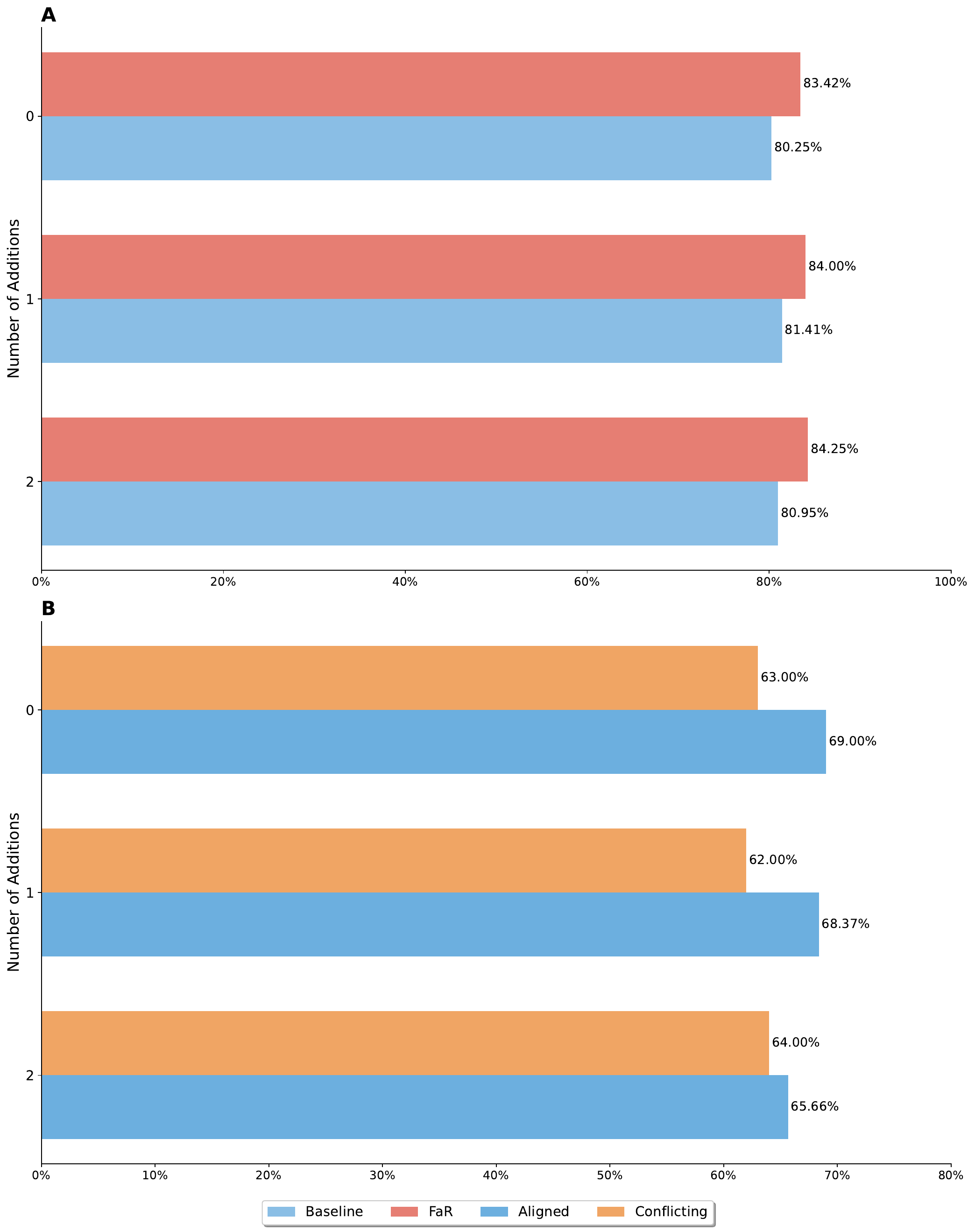}
    \captionsetup{width=1\linewidth}
    \caption{\textbf{Variation across number of additions.} Shows the performance across the number of additions (y-axis) that are done to all the schemas that have an implication statement either in their premise or in their conclusion (Section \ref{reasoning-depth-var}). The x-axis represents the accuracy values. \textbf{A} shows the comparison between \textit{FaR} and baseline approaches. We can see that \textit{FaR} improves performance in all the cases like in the case of 0 additions where performance improves from 80.22\% to 83.4\%.
    \textbf{B} shows the variation of Aligned and Conflicting scenarios across number of additions on valid data. We can see a pervasive performance gap between and Knowledge-Aligned and Knowledge-Conflicting data in all cases like in the case of 0 additions where the gap is about 6\%. However increasing the depth of reasoning doesn't result in a significant performance drop.}
    \label{fig:additions-comparison}
\end{figure}

\section{LLM Reasoning in Counterfactuals}
\label{sec:findings}

\subsection{Experimental Setup}
Our investigation evaluates 11 LLMs spanning different architectures, parameter scales, and training paradigms (see Appendix~\ref{app:models}). Evaluations were conducted across our \textit{CounterLogic} and five other diverse reasoning datasets: \textit{Hierarchical Syllogisms}, \textit{KNOT (Implicit \& Explicit)}, \textit{FOLIO}, and \textit{LogicBench} (see Table \ref{tab:dataset-comparison}). To ensure robust performance measures, we employ self-consistency checks accounting for the generation variability common in LLMs \cite{bonagiri2024sage}.
More deatils regarding the datasets and their evaluations is given in Appendix \ref{app:ground-truth-alignment}.

 Results presented in this sections establish a baseline using a standard CoT prompting strategy \cite{wei2023chainofthoughtpromptingelicitsreasoning} without our \textit{FaR} intervention, as this allows us to isolate and quantify the interference of parametric knowledge on logical deduction. 


\subsection{Consistent Performance Gap}

As shown in Figure~\ref{fig:combined-gap-analysis}, our evaluation revealed a pervasive performance gap across all models and tasks. When reasoning in counterfactual scenarios, where the logical context conflicts with their parametric knowledge, model accuracy plummets by an average of 14\%. Models achieve a high accuracy of 72\% on average for knowledge-aligned problems, but this value drops to just 58\% for counterfactual problems (Table \ref{tab:model_performance_valid_base}). Notably, this failure occurs even when models are explicitly instructed to reason based solely on the provided premises and ignore factual correctness (see Appendix~\ref{app:prompting-stragies}). This highlights a challenge in managing cognitive conflict rather than a simple failure to follow instructions. Even the most capable models exhibit this gap; for instance, Qwen-2-72B's accuracy difference is 20\% in the baseline setup. This suggests the interference from parametric knowledge is a fundamental challenge for current LLM reasoning. 


\subsection{Deconstructing the Failure}
\label{sec:deconstruct}


Is this performance degradation a superficial artifact of problem complexity or specific logical forms? We leveraged the controlled design of the \textit{CounterLogic} benchmark to investigate this, finding that the failure is more fundamental, affecting reasoning regardless of depth or logical structure.

A common hypothesis \cite{wu_reasoning_2024} is that models falter simply because counterfactual problems are more complex. Our analysis, however, refutes this by showing that the performance gap between knowledge-aligned and counterfactual examples is largely independent of the reasoning depth. As shown in Figure~\ref{fig:additions-comparison}, the core issue persists even as we increase the reasoning depth by introducing additional reasoning steps to the logical chain. We therefore hypothesize that the failure is not caused by models losing track of longer deductions, but rather by the initial, unresolved conflict with their parametric knowledge.

\label{sec:schema_variation}

Another possibility is that the performance drop is isolated to specific, difficult logical rules. Our results, shown in Figure 5(B), suggest a more nuanced reality. While the failure is a universal phenomenon across all 9 logical schemas in \textit{CounterLogic}, the degree of degradation varies dramatically, revealing critical vulnerabilities.

On simple, affirmative logical forms like Modus Ponens (MP) and Hypothetical Syllogism (HS), models are not only accurate but also more robust; the performance gap between knowledge-aligned and conflicting scenarios is minimal. However, this stability shatters when schemas introduce negation or greater structural complexity. On these tasks, the performance gap widens significantly, indicating that the model's parametric knowledge is interfering more strongly. For instance, the gap soars to over 30 percentage points for Disjunctive Syllogism (DS) and Destructive Dilemma (DD), both of which require reasoning over negated propositions. This suggests that logical complexity and negation don't just make the task harder—they specifically amplify the model's inability to inhibit its ingrained knowledge.

This systemic weakness in handling negation is also reflected in overall performance. Schemas involving negation have a baseline accuracy of only 73\%, 11\% lower than the 84\% achieved on those without (Figure 5(A)), aligning with findings that negation remains a core challenge for LLMs \cite{vrabcová2025negationpinkelephantlarge}. 

Ultimately, our analysis shows the failure is not confined to one type of logic. Instead, it is a foundational issue where the interference from parametric knowledge is severely exacerbated by structural complexity and the presence of negation.

\begin{figure}
    \centering
    \includegraphics[width=1\linewidth]{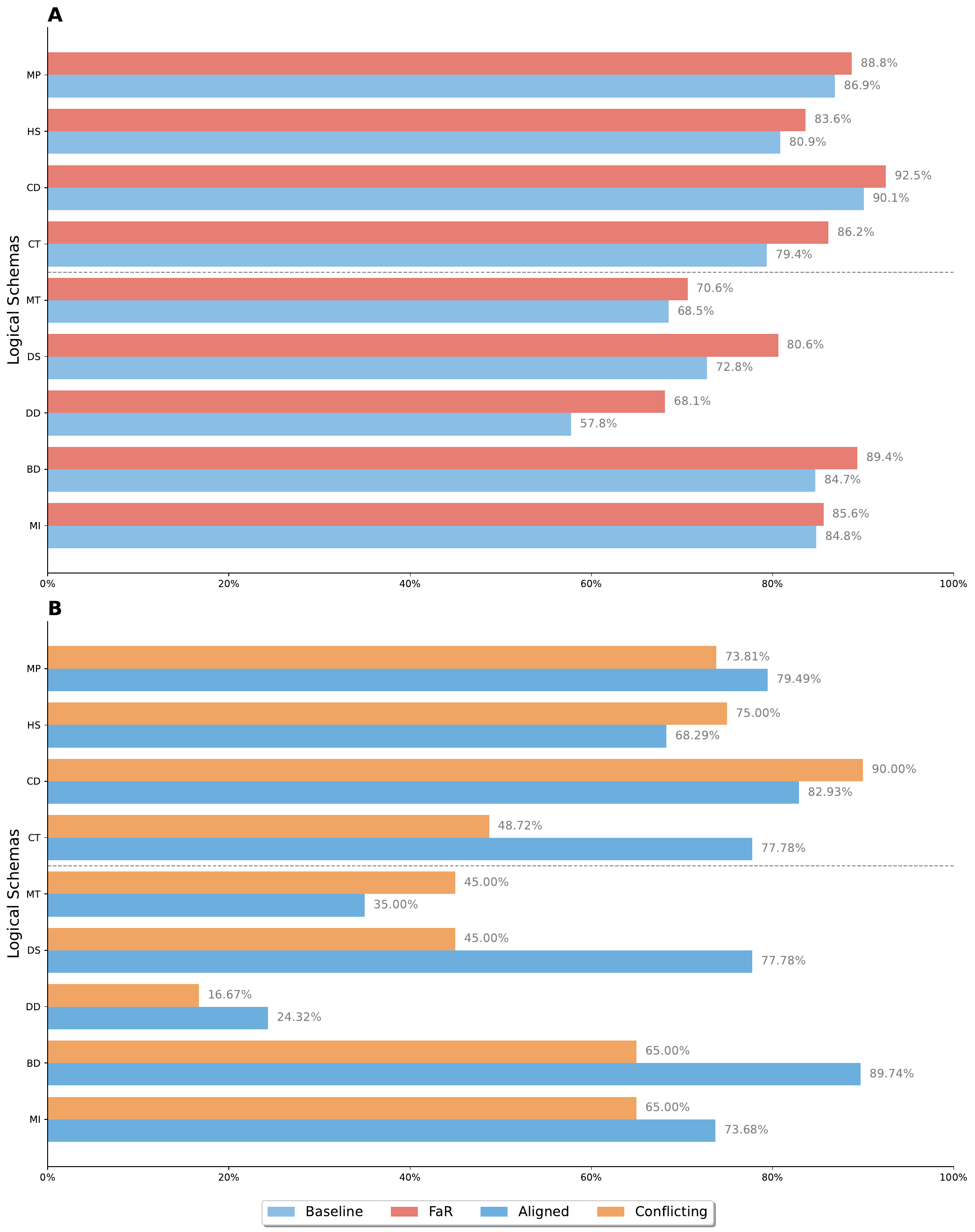}
    \captionsetup{width=1\linewidth}
    \caption{\textbf{Variation across logical schemas.} We can see the logical schemas (Table \ref{tab:logic-forms}) on the y-axis and the accuracies on the x-axis. Dotted line separates the schemas with (below) and without (top) negations. \textbf{A} shows the comparison between the \textit{FaR} and baseline accuracies. Pink bars represent \textit{FaR} method and blue bars represent the baseline. We can see that \textit{FaR} works across all schemas and improvement is more for the schemas that are harder (lower initial scores) like schema DD where the performance improves by about 11\%. We can also see how LLMs perform noticeably worse on tasks requiring understanding of negations with average performance dropping from 84.29\% above the line to 73.72\% below the line on the baseline method.
    \textbf{B} shows the performance difference of Knowledge-Aligned and Knowledge-Conflicting scenarios for valid datapoints. The blue bars represent aligned accuracies and the orange bars represent the conflicting accuracies. We see can see a clear performance gap between aligned and conflicting datapoints most noticeable in schema DS where the gap is about 32\%.}
    \label{fig:schema_eval}
\end{figure}
\section{Flag \& Reason (FaR): A MetaCognitive Intervention}
\label{sec:method_and_results}

Findings in Section~\ref{sec:findings} reveal a critical vulnerability in LLM reasoning: a systematic failure not of logic, but of cognitive management. In this section, we articulate our hypothesis for this failure, introduce the \textit{FaR} intervention designed to mitigate it, and present the empirical results that validate its effectiveness.

\subsection{The Belief Inhibition Hypothesis} 

The consistent performance gap between knowledge-aligned and counterfactual scenarios suggests that the primary issue is not a flaw in the
LLMs’ capacity for logical operations. Instead, we hypothesize that they suffer from a failure of \textit{belief inhibition}: an inability to suppress ingrained parametric knowledge when it contradicts the premises of a given task, leading to an unresolved cognitive conflict \cite{cognitivedissonance} that ultimately derails the deductive process.

This phenomenon closely mirrors a well-known cognitive bias in humans known as ``belief bias'', where arguments are often judged on the believability of their conclusion rather than the logical validity \cite{markovits_belief-bias_1989}. Humans, however, can overcome this bias through metacognition—the ability to reflect on one's own thinking. This involves consciously recognizing a conflict between a belief and a premise and choosing to set that belief aside to follow the rules of logic \cite{fletcher2012metacognition,Ma2025BeliefBias,wang2024metacognitivepromptingimprovesunderstanding}.

\subsection{The Flag \& Reason (FaR) Intervention}

Inspired by the human capability to recognize and set aside knowledge conflicts, we propose a simple yet powerful metacognitive intervention: \textbf{Flag \& Reason (FaR)}. Instead of asking the model to solve a logical problem at once, \textit{FaR} introduces a preliminary metacognitive step. The process, illustrated in Figure~\ref{fig:crownjewel}(B), consists of two phases:

\begin{enumerate}
    \item \textbf{Flag:} The model is prompted to first explicitly \textit{flag} whether a key premise or conclusion aligns with its parametric knowledge (i.e., whether it is factually believable).
    \item \textbf{Reason:} After this initial assessment, the model proceeds with the standard reasoning process, now primed by the ``Flag'' step to treat the premises as a self-contained logical context.
\end{enumerate}

The seperation of these two steps is critial. When we collapsed this process into a single prompt (asking the model to flag the conflict and reason in the same turn), the performance benefits disappeared entirely, yielding no improvement over the baseline (see Appendix~\ref{app:single-far}). This finding supports our central hypothesis: FaR works by inducing a form of \textbf{epistemic compartmentalization} \cite{thomas2013}. The initial ``Flag'' step forces the model to consciously acknowledge a boundary between its internal ``knowledge'' and the hypothetical ``rules'' of the task. This deliberate, separate, metacognitive act is what enables the model to suppress its belief bias and faithfully apply logic in the subsequent ``Reason'' step.

\subsection{Results}

\begin{figure}[t]
    \centering
    \includegraphics[width=\columnwidth]{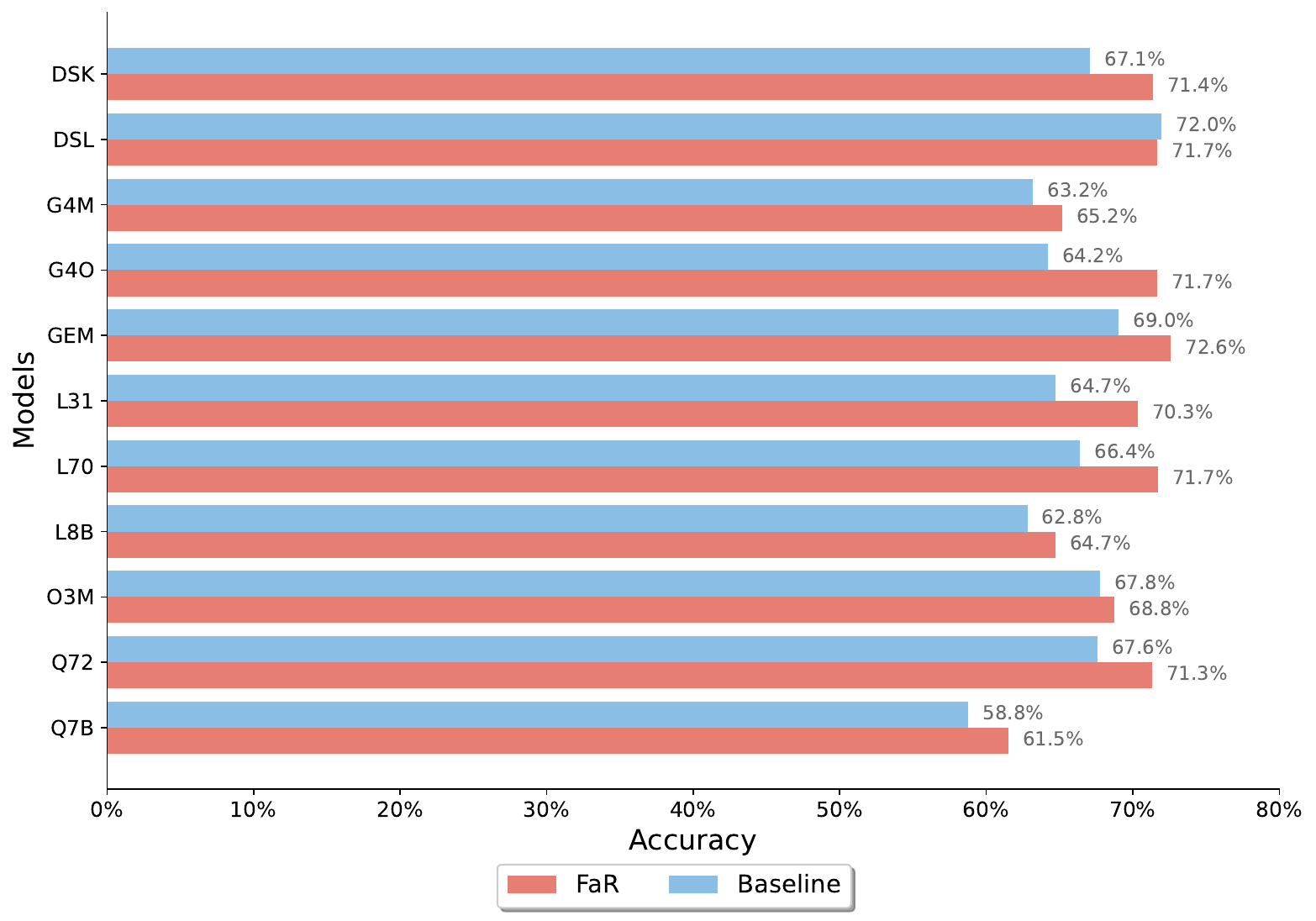} 
    
    \captionsetup{width=\columnwidth}
    
    \caption{%
    \textbf{Accuracy comparision between the baseline setup and our \textit{FaR} setup averaged across all the datasets.} 
    The x-axis represents accuracy and the y-axis represents the models. 
    Pink bars represent the \textit{FaR} method and the blue bars represent the baseline. 
    We can see that our method \textit{FaR} improves performance across all the models with the most noticeable improvement on GPT-4o (G4O) of about 7\%. 
    For Dataset wise results refer to figure \ref{fig:selfseg-dataset-comparison} in the appendix.%
    }
    \label{fig:selfseg-accuracy-comparison}
\end{figure}

Our experiments confirm that the \textit{FaR} intervention is highly effective. As detailed in Figure~\ref{fig:selfseg-accuracy-comparison}, \textit{FaR} narrows the average performance gap between knowledge-aligned and counterfactual scenarios from 14\% to just 7\%, while also increasing overall accuracy by an average of 4\% across all models and tasks (Table \ref{tab:model_performance}, \ref{tab:model-performance-valid-far}).

The most dramatic gains are observed on datasets like Hierarchical Syllogisms and KNOT. In contrast, tasks involving deeper chains of reasoning, such as FOLIO, showed less improvement, emphasizing the need for better conflict resolution strategies for complex narrative tasks \cite{han2024folionaturallanguagereasoning}. On our \textit{CounterLogic} benchmark specifically, \textit{FaR} increased overall valid accuracy by 5\%.  

Furthermore, this improvement is pervasive across all logical schemas in \textit{CounterLogic}. As shown in Figure \ref{fig:schema_eval}(A), \textit{FaR} provides a consistent boost, proving particularly effective for more complex schemas (e.g., Destructive Dilemma) that exhibit lower baseline performance. While schemas involving negation remain challenging for LLMs in general, \textit{FaR} improves average accuracy on these tasks from 73\% to 79\%.

\subsection{Thought Anchor Analysis}

To understand the mechanism behind these improvements, we analyzed the models' CoT generations using \textit{thought anchors} - critical reasoning steps that disproportionately influence the final outcome \cite{bogdan2025thoughtanchorsllmreasoning}. Our analysis focuses on the more human interpretable functional dynamics of the reasoning process, as opposed to typical mechanistic interpretabilty methods focusing on neuron/circuit level attributions. The full details of our analysis are presented in Appendix~\ref{app:thought-anchors}. 

Without the \textit{FaR} intervention, our analysis reveals that the reasoning process is heavily anchored in \textit{Fact Retrieval} and \textit{Uncertainty Management}. This observation provides strong evidence for our belief inhibition hypothesis: models appear to be sidetracked by retrieving conflicting parametric knowledge and then struggle to manage the resulting cognitive dissonance.
With the \textit{FaR} intervention, we observe a decisive repositioning of these thought anchors. The reasoning process becomes dominated by \textit{Plan Generation} and \textit{Active Computation}. Crucially, the model's reliance on retrieving external facts and expressing uncertainty is significantly diminished. This observed shift suggests a more stable and direct path for logical deduction. By prompting the model to first acknowledge the counterfactual nature of the premises, \textit{FaR} appears to effectively compartmentalize the conflict, allowing the model to proceed with the logical task with less interference from its ingrained knowledge. This fundamental change in the reasoning process offers a compelling explanation for the observed gains in accuracy.

\section{Conclusion and Future Work}
Our work confirms that LLMs exhibit a critical reasoning failure when logic conflicts with their parametric knowledge, showing a consistent 14\% accuracy drop in counterfactual scenarios, as measured on six reasoning datasets including our new \textit{CounterLogic} benchmark. We attribute this to a failure of \textit{belief inhibition}, a phenomenon similar to human cognitive biases. To address this, we introduced \textit{Flag \& Reason (FaR)}, a simple metacognitive intervention that not only halves this performance gap to 7\% but also boosts overall accuracy by 4\%. Our analysis of the models' reasoning chains reveals the mechanism for this success: \textit{FaR} helps the model to compartmentalize the knowledge conflict, thereby anchoring its deduction in the provided logical context instead of its contradictory parametric knowledge.

Future work could explore the internal model representations of these knowledge conflicts, extend metacognitive strategies to more complex, real-world reasoning domains, and develop methods to fundamentally resolve the tension between parametric memory and in-context logic, furthering the development of truly robust AI reasoners.

\section{Limitations}

While our study provides valuable insights, several limitations should be noted.
First, our \textit{CounterLogic} dataset cannot capture all real-world logical conflicts, as it focuses primarily on foundational structures like categorical syllogisms and propositional logic.
Second, our findings are specific to the models tested. The rapid pace of architectural innovation means future models may handle counterfactual information differently.
Third, our evaluation was limited to English and common-knowledge concepts. The interventions' effectiveness may vary across other languages, cultures, and specialized domains.
Fourth, parallels drawn to human cognition are conceptual frameworks and should not be interpreted as evidence of identical cognitive processes in LLMs.
Finally, using LLMs to synthetically generate queries may have introduced unnoticed inconsistencies or biases from the generator models themselves.

Despite these limitations, our findings demonstrate substantial improvements in counterfactual reasoning, suggesting that our core insights will remain relevant as models evolve.

\section{Acknowledgments}

This work was supported by the OpenAI Research Access Program which partially provided the necessary funding for API credits. We thank Hanuma, Sumit Kumar, Ameya Rathod, Vedant SP, Vaishnavi Shivkumar, Hari Shankar and other members from the Precog research group for their insightful discussions and valuable feedback. Manas Gaur, TK Prasad, and P. Kumaraguru provided valuable feedback during all phases of the project. 

\bibliographystyle{IEEEtran}
\bibliography{custom}

\begin{thebibliography}{10}
\providecommand{\url}[1]{#1}
\csname url@samestyle\endcsname
\providecommand{\newblock}{\relax}
\providecommand{\bibinfo}[2]{#2}
\providecommand{\BIBentrySTDinterwordspacing}{\spaceskip=0pt\relax}
\providecommand{\BIBentryALTinterwordstretchfactor}{4}
\providecommand{\BIBentryALTinterwordspacing}{\spaceskip=\fontdimen2\font plus
\BIBentryALTinterwordstretchfactor\fontdimen3\font minus
  \fontdimen4\font\relax}
\providecommand{\BIBforeignlanguage}[2]{{%
\expandafter\ifx\csname l@#1\endcsname\relax
\typeout{** WARNING: IEEEtran.bst: No hyphenation pattern has been}%
\typeout{** loaded for the language `#1'. Using the pattern for}%
\typeout{** the default language instead.}%
\else
\language=\csname l@#1\endcsname
\fi
#2}}
\providecommand{\BIBdecl}{\relax}
\BIBdecl

\bibitem{brown2020language}
\BIBentryALTinterwordspacing
T.~Brown, B.~Mann, N.~Ryder, M.~Subbiah, J.~D. Kaplan, P.~Dhariwal,
  A.~Neelakantan, P.~Shyam, G.~Sastry, A.~Askell, S.~Agarwal, A.~Herbert-Voss,
  G.~Krueger, T.~Henighan, R.~Child, A.~Ramesh, D.~Ziegler, J.~Wu, C.~Winter,
  C.~Hesse, M.~Chen, E.~Sigler, M.~Litwin, S.~Gray, B.~Chess, J.~Clark,
  C.~Berner, S.~McCandlish, A.~Radford, I.~Sutskever, and D.~Amodei, ``Language
  models are few-shot learners,'' \emph{Advances in Neural Information
  Processing Systems}, vol.~33, pp. 1877--1901, 2020. [Online]. Available:
  \url{https://proceedings.neurips.cc/paper_files/paper/2020/file/1457c0d6bfcb4967418bfb8ac142f64a-Paper.pdf}
\BIBentrySTDinterwordspacing

\bibitem{kojima2022large}
T.~Kojima, S.~S. Gu, M.~Reid, Y.~Matsuo, and Y.~Iwasawa, ``Large language
  models are zero-shot reasoners,'' \emph{arXiv preprint arXiv:2205.11916},
  2022.

\bibitem{schick2021small}
\BIBentryALTinterwordspacing
T.~Schick and H.~Sch{\"u}tze, ``It`s not just size that matters: Small language
  models are also few-shot learners,'' in \emph{Proceedings of the 2021
  Conference of the North American Chapter of the Association for Computational
  Linguistics: Human Language Technologies}, K.~Toutanova, A.~Rumshisky,
  L.~Zettlemoyer, D.~Hakkani-Tur, I.~Beltagy, S.~Bethard, R.~Cotterell,
  T.~Chakraborty, and Y.~Zhou, Eds.\hskip 1em plus 0.5em minus 0.4em\relax
  Online: Association for Computational Linguistics, Jun. 2021, pp. 2339--2352.
  [Online]. Available: \url{https://aclanthology.org/2021.naacl-main.185/}
\BIBentrySTDinterwordspacing

\bibitem{wei2023chainofthoughtpromptingelicitsreasoning}
\BIBentryALTinterwordspacing
J.~Wei, X.~Wang, D.~Schuurmans, M.~Bosma, B.~Ichter, F.~Xia, E.~Chi, Q.~Le, and
  D.~Zhou, ``Chain-of-thought prompting elicits reasoning in large language
  models,'' 2023. [Online]. Available: \url{https://arxiv.org/abs/2201.11903}
\BIBentrySTDinterwordspacing

\bibitem{zhou2023least}
D.~Zhou, N.~Schärli, L.~Hou, J.~Wei, N.~Scales, X.~Wang, D.~Schuurmans,
  C.~Cui, O.~Bousquet, Q.~Le, and E.~Chi, ``Least-to-most prompting enables
  complex reasoning in large language models,'' in \emph{International
  Conference on Learning Representations}, 2023.

\bibitem{patel2024multi}
\BIBentryALTinterwordspacing
N.~Patel, M.~Kulkarni, M.~Parmar, A.~Budhiraja, M.~Nakamura, N.~Varshney, and
  C.~Baral, ``Multi-logieval: Towards evaluating multi-step logical reasoning
  ability of large language models,'' in \emph{Proceedings of EMNLP}, 2024.
  [Online]. Available: \url{https://aclanthology.org/2024.emnlp-main.1160.pdf}
\BIBentrySTDinterwordspacing

\bibitem{dasgupta_language_2024}
\BIBentryALTinterwordspacing
I.~Dasgupta, A.~K. Lampinen, S.~C.~Y. Chan, H.~R. Sheahan, A.~Creswell,
  D.~Kumaran, J.~L. McClelland, and F.~Hill, ``Language models show human-like
  content effects on reasoning tasks,'' Jul. 2024, arXiv:2207.07051 [cs].
  [Online]. Available: \url{http://arxiv.org/abs/2207.07051}
\BIBentrySTDinterwordspacing

\bibitem{jin2024tug}
Z.~Jin, P.~Cao, Y.~Chen, K.~Liu, X.~Jiang, J.~Xu, L.~Qiuxia, and J.~Zhao,
  ``Tug-of-war between knowledge: Exploring and resolving knowledge conflicts
  in retrieval-augmented language models,'' in \emph{Proceedings of
  LREC-COLING}, 2024, pp. 10\,142--10\,151.

\bibitem{lampinen2024language}
A.~K. Lampinen, I.~Dasgupta, S.~C. Chan, H.~R. Sheahan, A.~Creswell,
  D.~Kumaran, J.~L. McClelland, and F.~Hill, ``Language models, like humans,
  show content effects on reasoning tasks,'' \emph{PNAS Nexus}, vol.~3, no.~7,
  p. pgae233, 2024.

\bibitem{su2024conflictbank}
\BIBentryALTinterwordspacing
Z.~Su, J.~Zhang, X.~Qu, T.~Zhu, Y.~Li, J.~Sun, J.~Li, M.~Zhang, and Y.~Cheng,
  ``Conflictbank: A benchmark for evaluating the influence of knowledge
  conflicts in llm,'' 2024. [Online]. Available:
  \url{https://arxiv.org/abs/2408.12076}
\BIBentrySTDinterwordspacing

\bibitem{wang_resolving_2024}
\BIBentryALTinterwordspacing
Y.~Wang, S.~Feng, H.~Wang, W.~Shi, V.~Balachandran, T.~He, and Y.~Tsvetkov,
  ``Resolving {Knowledge} {Conflicts} in {Large} {Language} {Models},'' Oct.
  2024, arXiv:2310.00935 [cs]. [Online]. Available:
  \url{http://arxiv.org/abs/2310.00935}
\BIBentrySTDinterwordspacing

\bibitem{wu_reasoning_2024}
\BIBentryALTinterwordspacing
Z.~Wu, L.~Qiu, A.~Ross, E.~Akyürek, B.~Chen, B.~Wang, N.~Kim, J.~Andreas, and
  Y.~Kim, ``Reasoning or {Reciting}? {Exploring} the {Capabilities} and
  {Limitations} of {Language} {Models} {Through} {Counterfactual} {Tasks},''
  Mar. 2024, arXiv:2307.02477 [cs]. [Online]. Available:
  \url{http://arxiv.org/abs/2307.02477}
\BIBentrySTDinterwordspacing

\bibitem{xu_knowledge_2024}
\BIBentryALTinterwordspacing
R.~Xu, Z.~Qi, Z.~Guo, C.~Wang, H.~Wang, Y.~Zhang, and W.~Xu, ``Knowledge
  {Conflicts} for {LLMs}: {A} {Survey},'' in \emph{Proceedings of the 2024
  {Conference} on {Empirical} {Methods} in {Natural} {Language} {Processing}},
  Y.~Al-Onaizan, M.~Bansal, and Y.-N. Chen, Eds.\hskip 1em plus 0.5em minus
  0.4em\relax Miami, Florida, USA: Association for Computational Linguistics,
  Nov. 2024, pp. 8541--8565. [Online]. Available:
  \url{https://aclanthology.org/2024.emnlp-main.486/}
\BIBentrySTDinterwordspacing

\bibitem{pearl2018book}
\BIBentryALTinterwordspacing
J.~Pearl and D.~Mackenzie, \emph{The Book of Why: The New Science of Cause and
  Effect}.\hskip 1em plus 0.5em minus 0.4em\relax Basic Books, 2018. [Online].
  Available: \url{https://en.wikipedia.org/wiki/The_Book_of_Why}
\BIBentrySTDinterwordspacing

\bibitem{zhang2024bridging}
\BIBentryALTinterwordspacing
Y.~Zhang, W.~Wang, X.~Liu, Y.~Chen, and Z.~Li, ``Bridging the gap between llms
  and human intentions,'' \emph{arXiv preprint arXiv:2502.09101}, 2024.
  [Online]. Available: \url{https://arxiv.org/abs/2502.09101}
\BIBentrySTDinterwordspacing

\bibitem{lewis-2024-counterfactual}
M.~Lewis and M.~Mitchell, ``Using counterfactual tasks to evaluate the
  generality of analogical reasoning in large language models,'' \emph{arXiv
  preprint arXiv:2402.08955}, 2024.

\bibitem{liu_untangle_2024}
\BIBentryALTinterwordspacing
Y.~Liu, Z.~Yao, X.~Lv, Y.~Fan, S.~Cao, J.~Yu, L.~Hou, and J.~Li, ``Untangle the
  {KNOT}: {Interweaving} {Conflicting} {Knowledge} and {Reasoning} {Skills} in
  {Large} {Language} {Models},'' Apr. 2024, arXiv:2404.03577 [cs]. [Online].
  Available: \url{http://arxiv.org/abs/2404.03577}
\BIBentrySTDinterwordspacing

\bibitem{kaushik-etal-2020-learning}
D.~Kaushik, E.~Hovy, and Z.~C. Lipton, ``Learning the difference that makes a
  difference with counterfactually-augmented data,'' in \emph{Proceedings of
  the International Conference on Learning Representations}, 2020.

\bibitem{mccoy-etal-2019-right}
R.~T. McCoy, E.~Pavlick, and T.~Linzen, ``Right for the wrong reasons:
  Diagnosing syntactic heuristics in natural language inference,'' in
  \emph{Proceedings of the 57th Annual Meeting of the Association for
  Computational Linguistics}, Florence, Italy, 2019, pp. 3428--3448.

\bibitem{xie_adaptive_2024}
\BIBentryALTinterwordspacing
J.~Xie, K.~Zhang, J.~Chen, R.~Lou, and Y.~Su, ``Adaptive {Chameleon} or
  {Stubborn} {Sloth}: {Revealing} the {Behavior} of {Large} {Language} {Models}
  in {Knowledge} {Conflicts},'' Feb. 2024, arXiv:2305.13300 [cs]. [Online].
  Available: \url{http://arxiv.org/abs/2305.13300}
\BIBentrySTDinterwordspacing

\bibitem{fletcher2012metacognition}
L.~Fletcher and P.~Carruthers, ``Metacognition and reasoning,''
  \emph{Philosophical Transactions of the Royal Society B: Biological
  Sciences}, vol. 367, no. 1594, pp. 1366--1378, 2012.

\bibitem{parmar2024logicbench}
\BIBentryALTinterwordspacing
M.~Parmar, N.~Patel, N.~Varshney, M.~Nakamura, M.~Luo, S.~Mashetty, A.~Mitra,
  and C.~Baral, ``Logicbench: Towards systematic evaluation of logical
  reasoning ability of large language models,'' 2024. [Online]. Available:
  \url{https://arxiv.org/abs/2404.15522}
\BIBentrySTDinterwordspacing

\bibitem{han2024folionaturallanguagereasoning}
\BIBentryALTinterwordspacing
S.~Han, H.~Schoelkopf, Y.~Zhao, Z.~Qi, M.~Riddell, W.~Zhou, J.~Coady, D.~Peng,
  Y.~Qiao, L.~Benson, L.~Sun, A.~Wardle-Solano, H.~Szabo, E.~Zubova,
  M.~Burtell, J.~Fan, Y.~Liu, B.~Wong, M.~Sailor, A.~Ni, L.~Nan, J.~Kasai,
  T.~Yu, R.~Zhang, A.~R. Fabbri, W.~Kryscinski, S.~Yavuz, Y.~Liu, X.~V. Lin,
  S.~Joty, Y.~Zhou, C.~Xiong, R.~Ying, A.~Cohan, and D.~Radev, ``Folio: Natural
  language reasoning with first-order logic,'' 2024. [Online]. Available:
  \url{https://arxiv.org/abs/2209.00840}
\BIBentrySTDinterwordspacing

\bibitem{bertolazzi2024systematicanalysislargelanguage}
\BIBentryALTinterwordspacing
L.~Bertolazzi, A.~Gatt, and R.~Bernardi, ``A systematic analysis of large
  language models as soft reasoners: The case of syllogistic inferences,''
  2024. [Online]. Available: \url{https://arxiv.org/abs/2406.11341}
\BIBentrySTDinterwordspacing

\bibitem{yao2023tree}
\BIBentryALTinterwordspacing
N.~Shinn, S.~Yao, K.~Zhao, D.~Yu, E.~Zhao, D.~Zhao, and D.~Radev, ``Tree of
  thoughts: Deliberate problem solving with large language models,'' 2023.
  [Online]. Available: \url{https://arxiv.org/abs/2305.10601}
\BIBentrySTDinterwordspacing

\bibitem{clark2020transformers}
\BIBentryALTinterwordspacing
E.~Clark, O.~Tafjord, K.~Richardson, A.~Sabharwal, and H.~Hajishirzi,
  ``Transformers as soft reasoners over language,'' in \emph{Proceedings of the
  58th Annual Meeting of the Association for Computational Linguistics (ACL)},
  2020, pp. 3882--3894. [Online]. Available:
  \url{https://aclanthology.org/2020.acl-main.358}
\BIBentrySTDinterwordspacing

\bibitem{eisape_systematic_2024}
\BIBentryALTinterwordspacing
T.~Eisape, M.~H. Tessler, I.~Dasgupta, F.~Sha, S.~v. Steenkiste, and T.~Linzen,
  ``A {Systematic} {Comparison} of {Syllogistic} {Reasoning} in {Humans} and
  {Language} {Models},'' Apr. 2024, arXiv:2311.00445 [cs]. [Online]. Available:
  \url{http://arxiv.org/abs/2311.00445}
\BIBentrySTDinterwordspacing

\bibitem{singhprobing}
S.~Singh, S.~Goel, S.~Vaduguru, and P.~Kumaraguru, ``Probing negation in
  language models,'' 2024.

\bibitem{chen2025counterbench}
\BIBentryALTinterwordspacing
Y.~Chen, V.~K. Singh, J.~Ma, and R.~Tang, ``Counterbench: A benchmark for
  counterfactual reasoning in large language models,'' \emph{arXiv preprint
  arXiv:2502.11008}, 2025. [Online]. Available:
  \url{https://arxiv.org/abs/2502.11008}
\BIBentrySTDinterwordspacing

\bibitem{estermann2025reasoning}
\BIBentryALTinterwordspacing
B.~Estermann, L.~A. Lanzend{\"o}rfer, and R.~Wattenhofer, ``Reasoning effort
  and problem complexity: A scaling analysis in large language models,''
  \emph{arXiv preprint arXiv:2503.15113}, 2025. [Online]. Available:
  \url{https://arxiv.org/abs/2503.15113}
\BIBentrySTDinterwordspacing

\bibitem{xu2024faithful}
J.~Xu, H.~Fei, L.~Pan, Q.~Liu, M.~Lee, and W.~Hsu, ``Faithful logical reasoning
  via symbolic chain-of-thought,'' \emph{Annual Meeting of the Association for
  Computational Linguistics}, 2024.

\bibitem{toroghi2024verifiable}
\BIBentryALTinterwordspacing
A.~Toroghi, W.~Guo, and S.~Sanner, ``Right for right reasons: Large language
  models for verifiable commonsense knowledge graph question answering,'' in
  \emph{Proceedings of the 2024 Conference on Empirical Methods in Natural
  Language Processing}, 2024. [Online]. Available:
  \url{https://arxiv.org/abs/2403.01390}
\BIBentrySTDinterwordspacing

\bibitem{vacareanu2024general}
\BIBentryALTinterwordspacing
R.~Vacareanu and M.~Ballesteros, ``General purpose verification for chain of
  thought prompting,'' \emph{arXiv preprint arXiv:2405.00204}, 2024. [Online].
  Available: \url{https://arxiv.org/abs/2405.00204}
\BIBentrySTDinterwordspacing

\bibitem{petroni_language_2019}
\BIBentryALTinterwordspacing
F.~Petroni, T.~Rocktäschel, S.~Riedel, P.~Lewis, A.~Bakhtin, Y.~Wu, and
  A.~Miller, ``Language {Models} as {Knowledge} {Bases}?'' in \emph{Proceedings
  of the 2019 {Conference} on {Empirical} {Methods} in {Natural} {Language}
  {Processing} and the 9th {International} {Joint} {Conference} on {Natural}
  {Language} {Processing} ({EMNLP}-{IJCNLP})}, K.~Inui, J.~Jiang, V.~Ng, and
  X.~Wan, Eds.\hskip 1em plus 0.5em minus 0.4em\relax Hong Kong, China:
  Association for Computational Linguistics, Nov. 2019, pp. 2463--2473.
  [Online]. Available: \url{https://aclanthology.org/D19-1250/}
\BIBentrySTDinterwordspacing

\bibitem{roberts_how_2020}
\BIBentryALTinterwordspacing
A.~Roberts, C.~Raffel, and N.~Shazeer, ``How {Much} {Knowledge} {Can} {You}
  {Pack} {Into} the {Parameters} of a {Language} {Model}?'' Oct. 2020,
  arXiv:2002.08910 [cs]. [Online]. Available:
  \url{http://arxiv.org/abs/2002.08910}
\BIBentrySTDinterwordspacing

\bibitem{longpre_entity-based_2021}
\BIBentryALTinterwordspacing
S.~Longpre, K.~Perisetla, A.~Chen, N.~Ramesh, C.~DuBois, and S.~Singh,
  ``Entity-based knowledge conflicts in question answering,'' in
  \emph{Proceedings of the 2021 Conference on Empirical Methods in Natural
  Language Processing}.\hskip 1em plus 0.5em minus 0.4em\relax Association for
  Computational Linguistics, 2021, pp. 7052--7063. [Online]. Available:
  \url{https://aclanthology.org/2021.emnlp-main.565/}
\BIBentrySTDinterwordspacing

\bibitem{Wangetal2023}
\BIBentryALTinterwordspacing
C.~Wang, X.~Liu, Y.~Yue, X.~Tang, T.~Zhang, C.~Jiayang, Y.~Yao, W.~Gao, X.~Hu,
  Z.~Qi, Y.~Wang, L.~Yang, J.~Wang, X.~Xie, Z.~Zhang, and Y.~Zhang, ``Survey on
  factuality in large language models: Knowledge, retrieval and
  domain-specificity,'' 2023. [Online]. Available:
  \url{https://arxiv.org/abs/2310.07521}
\BIBentrySTDinterwordspacing

\bibitem{lin2025generative}
\BIBentryALTinterwordspacing
H.~Lin, X.~Wang, R.~Yan, B.~Huang, H.~Ye, J.~Zhu, Z.~Wang, J.~Zou, J.~Ma, and
  Y.~Liang, ``Generative reasoning with large language models,'' 2025.
  [Online]. Available: \url{https://arxiv.org/abs/2504.02810}
\BIBentrySTDinterwordspacing

\bibitem{chen2023discodistillingcounterfactualslarge}
\BIBentryALTinterwordspacing
Z.~Chen, Q.~Gao, A.~Bosselut, A.~Sabharwal, and K.~Richardson, ``Disco:
  Distilling counterfactuals with large language models,'' 2023. [Online].
  Available: \url{https://arxiv.org/abs/2212.10534}
\BIBentrySTDinterwordspacing

\bibitem{neeman2022disentqa}
\BIBentryALTinterwordspacing
E.~Neeman, R.~Aharoni, O.~Honovich, L.~Choshen, I.~Szpektor, and O.~Abend,
  ``Disentqa: Disentangled question answering,'' in \emph{Proceedings of the
  61st Annual Meeting of the Association for Computational Linguistics (Volume
  1: Long Papers)}, A.~Rogers, J.~Boyd-Graber, and N.~Okazaki, Eds.\hskip 1em
  plus 0.5em minus 0.4em\relax Toronto, Canada: Association for Computational
  Linguistics, Jul. 2023, pp. 10\,056--10\,070. [Online]. Available:
  \url{https://aclanthology.org/2023.acl-long.559/}
\BIBentrySTDinterwordspacing

\bibitem{markovits_belief-bias_1989}
\BIBentryALTinterwordspacing
H.~Markovits and G.~Nantel, ``\BIBforeignlanguage{en}{The belief-bias effect in
  the production and evaluation of logical conclusions},''
  \emph{\BIBforeignlanguage{en}{Memory \& Cognition}}, vol.~17, no.~1, pp.
  11--17, Jan. 1989. [Online]. Available:
  \url{https://doi.org/10.3758/BF03199552}
\BIBentrySTDinterwordspacing

\bibitem{kunda1990case}
\BIBentryALTinterwordspacing
Z.~Kunda, ``The case for motivated reasoning,'' \emph{Psychological Bulletin},
  vol. 108, no.~3, pp. 480--498, 1990. [Online]. Available:
  \url{https://doi.org/10.1037/0033-2909.108.3.480}
\BIBentrySTDinterwordspacing

\bibitem{trippas2014fluency}
D.~Trippas, S.~Handley, and M.~Verde, ``Fluency and belief bias in deductive
  reasoning: new indices for old effects,'' \emph{Frontiers in Psychology},
  vol.~5, p. 631, 06 2014.

\bibitem{macmillan-scott2024irrationality}
\BIBentryALTinterwordspacing
O.~Macmillan-Scott and M.~Musolesi, ``(ir)rationality and cognitive biases in
  large language models,'' \emph{Royal Society Open Science}, vol.~11, no.~3,
  p. 240255, 2024. [Online]. Available:
  \url{https://doi.org/10.1098/rsos.240255}
\BIBentrySTDinterwordspacing

\bibitem{douven2018conditionals}
\BIBentryALTinterwordspacing
I.~Douven, S.~Elqayam, and H.~Singmann, ``Conditionals and inferential
  connections: Toward a new semantics,'' \emph{Cognition}, vol. 178, pp.
  31--45, 2018. [Online]. Available:
  \url{https://doi.org/10.1016/j.cognition.2018.05.005}
\BIBentrySTDinterwordspacing

\bibitem{zhou_navigating_2023}
\BIBentryALTinterwordspacing
K.~Zhou, D.~Jurafsky, and T.~Hashimoto, ``Navigating the {Grey} {Area}: {How}
  {Expressions} of {Uncertainty} and {Overconfidence} {Affect} {Language}
  {Models},'' Nov. 2023, arXiv:2302.13439 [cs]. [Online]. Available:
  \url{http://arxiv.org/abs/2302.13439}
\BIBentrySTDinterwordspacing

\bibitem{chen_say_2023}
\BIBentryALTinterwordspacing
J.~Chen, W.~Shi, Z.~Fu, S.~Cheng, L.~Li, and Y.~Xiao, ``Say {What} {You}
  {Mean}! {Large} {Language} {Models} {Speak} {Too} {Positively} about
  {Negative} {Commonsense} {Knowledge},'' in \emph{Proceedings of the 61st
  {Annual} {Meeting} of the {Association} for {Computational} {Linguistics}
  ({Volume} 1: {Long} {Papers})}, A.~Rogers, J.~Boyd-Graber, and N.~Okazaki,
  Eds.\hskip 1em plus 0.5em minus 0.4em\relax Toronto, Canada: Association for
  Computational Linguistics, Jul. 2023, pp. 9890--9908. [Online]. Available:
  \url{https://aclanthology.org/2023.acl-long.550/}
\BIBentrySTDinterwordspacing

\bibitem{xie2024memorization}
\BIBentryALTinterwordspacing
C.~Xie, Y.~Huang, C.~Zhang, D.~Yu, X.~Chen, B.~Y. Lin, B.~Li, B.~Ghazi, and
  R.~Kumar, ``On memorization of large language models in logical reasoning,''
  \emph{arXiv preprint arXiv:2410.23123}, 2024. [Online]. Available:
  \url{https://arxiv.org/abs/2410.23123}
\BIBentrySTDinterwordspacing

\bibitem{bonagiri2024sage}
V.~K. Bonagiri, S.~Vennam, P.~Govil, P.~Kumaraguru, and M.~Gaur, ``Sage:
  Evaluating moral consistency in large language models,'' \emph{arXiv preprint
  arXiv:2402.13709}, 2024.

\bibitem{vrabcová2025negationpinkelephantlarge}
\BIBentryALTinterwordspacing
T.~Vrabcová, M.~Kadlčík, P.~Sojka, M.~Štefánik, and M.~Spiegel,
  ``Negation: A pink elephant in the large language models' room?'' 2025.
  [Online]. Available: \url{https://arxiv.org/abs/2503.22395}
\BIBentrySTDinterwordspacing

\bibitem{cognitivedissonance}
\BIBentryALTinterwordspacing
E.~Harmon-Jones, \emph{Cognitive Dissonance: Reexamining a Pivotal Theory in
  Psychology}, 2nd~ed.\hskip 1em plus 0.5em minus 0.4em\relax American
  Psychological Association, 2019. [Online]. Available:
  \url{http://www.jstor.org/stable/j.ctv1chs6tk}
\BIBentrySTDinterwordspacing

\bibitem{Ma2025BeliefBias}
J.~Ma, W.~Lv, and X.~Ren, ``Neural correlates of belief-bias reasoning as
  predictors of critical thinking: Evidence from an fnirs study,''
  \emph{Journal of Intelligence}, vol.~13, no.~9, p. 106, Aug. 2025.

\bibitem{wang2024metacognitivepromptingimprovesunderstanding}
\BIBentryALTinterwordspacing
Y.~Wang and Y.~Zhao, ``Metacognitive prompting improves understanding in large
  language models,'' 2024. [Online]. Available:
  \url{https://arxiv.org/abs/2308.05342}
\BIBentrySTDinterwordspacing

\bibitem{thomas2013}
J.~Thomas, C.~Ditzfeld, and C.~Showers, ``Compartmentalization: A window on the
  defensive self 1,'' \emph{Social and Personality Psychology Compass},
  vol.~10, pp. 719--7\,311\,111, 10 2013.

\bibitem{bogdan2025thoughtanchorsllmreasoning}
\BIBentryALTinterwordspacing
P.~C. Bogdan, U.~Macar, N.~Nanda, and A.~Conmy, ``Thought anchors: Which llm
  reasoning steps matter?'' 2025. [Online]. Available:
  \url{https://arxiv.org/abs/2506.19143}
\BIBentrySTDinterwordspacing

\bibitem{10.1145/3194770.3194776}
\BIBentryALTinterwordspacing
S.~Verma and J.~Rubin, ``Fairness definitions explained,'' in \emph{Proceedings
  of the International Workshop on Software Fairness}, ser. FairWare '18.\hskip
  1em plus 0.5em minus 0.4em\relax New York, NY, USA: Association for Computing
  Machinery, 2018, p. 1–7. [Online]. Available:
  \url{https://doi.org/10.1145/3194770.3194776}
\BIBentrySTDinterwordspacing

\bibitem{10273594}
L.~Huang, Y.~Wu, and N.~Tempini, ``A knowledge flow empowered cognitive
  framework for decision making with task-agnostic data regulation,''
  \emph{IEEE Transactions on Artificial Intelligence}, vol.~5, no.~5, pp.
  2304--2318, 2024.

\bibitem{hong2025impliesbcircuitanalysis}
\BIBentryALTinterwordspacing
G.~Z. Hong, N.~Dikkala, E.~Luo, C.~Rashtchian, X.~Wang, and R.~Panigrahy, ``A
  implies b: Circuit analysis in llms for propositional logical reasoning,''
  2025. [Online]. Available: \url{https://arxiv.org/abs/2411.04105}
\BIBentrySTDinterwordspacing

\bibitem{KUGARAJEEVAN2025125381}
\BIBentryALTinterwordspacing
J.~Kugarajeevan, K.~Thanikasalam, A.~Ramanan, and S.~Fernando, ``Selective
  information flow for transformer tracking,'' \emph{Expert Systems with
  Applications}, vol. 259, p. 125381, 2025. [Online]. Available:
  \url{https://www.sciencedirect.com/science/article/pii/S0957417424022486}
\BIBentrySTDinterwordspacing

\bibitem{doi:https://doi.org/10.1002/9781405165518.wbeosc058.pub2}
\BIBentryALTinterwordspacing
M.~K. Miller, J.~D. Clark, and A.~Jehle, \emph{Cognitive Dissonance Theory
  (Festinger)}.\hskip 1em plus 0.5em minus 0.4em\relax John Wiley \& Sons, Ltd,
  2015. [Online]. Available:
  \url{https://onlinelibrary.wiley.com/doi/abs/10.1002/9781405165518.wbeosc058.pub2}
\BIBentrySTDinterwordspacing

\bibitem{unknown}
L.~Griffin, B.~Kleinberg, M.~Mozes, K.~Mai, M.~Vau, M.~Caldwell, and
  A.~Marvor-Parker, ``Susceptibility to influence of large language models,''
  03 2023.

\bibitem{mundler2024selfcontradictory}
\BIBentryALTinterwordspacing
N.~M{\"u}ndler, J.~He, S.~Jenko, and M.~Vechev, ``Self-contradictory
  hallucinations of large language models: Evaluation, detection and
  mitigation,'' in \emph{The Twelfth International Conference on Learning
  Representations}, 2024. [Online]. Available:
  \url{https://openreview.net/forum?id=EmQSOi1X2f}
\BIBentrySTDinterwordspacing

\bibitem{rozner2024knowledgeeditinglanguagemodels}
\BIBentryALTinterwordspacing
A.~Rozner, B.~Battash, L.~Wolf, and O.~Lindenbaum, ``Knowledge editing in
  language models via adapted direct preference optimization,'' 2024. [Online].
  Available: \url{https://arxiv.org/abs/2406.09920}
\BIBentrySTDinterwordspacing

\bibitem{gutiérrez2025hipporagneurobiologicallyinspiredlongterm}
\BIBentryALTinterwordspacing
B.~J. Gutiérrez, Y.~Shu, Y.~Gu, M.~Yasunaga, and Y.~Su, ``Hipporag:
  Neurobiologically inspired long-term memory for large language models,''
  2025. [Online]. Available: \url{https://arxiv.org/abs/2405.14831}
\BIBentrySTDinterwordspacing

\bibitem{Huang_2025}
\BIBentryALTinterwordspacing
L.~Huang, W.~Yu, W.~Ma, W.~Zhong, Z.~Feng, H.~Wang, Q.~Chen, W.~Peng, X.~Feng,
  B.~Qin, and T.~Liu, ``A survey on hallucination in large language models:
  Principles, taxonomy, challenges, and open questions,'' \emph{ACM
  Transactions on Information Systems}, vol.~43, no.~2, p. 1–55, Jan. 2025.
  [Online]. Available: \url{http://dx.doi.org/10.1145/3703155}
\BIBentrySTDinterwordspacing

\bibitem{zhang2024contrasolverselfalignmentlanguagemodels}
\BIBentryALTinterwordspacing
X.~Zhang, X.~Yin, and X.~Wan, ``Contrasolver: Self-alignment of language models
  by resolving internal preference contradictions,'' 2024. [Online]. Available:
  \url{https://arxiv.org/abs/2406.08842}
\BIBentrySTDinterwordspacing

\bibitem{venhoff2025understandingreasoningthinkinglanguage}
\BIBentryALTinterwordspacing
C.~Venhoff, I.~Arcuschin, P.~Torr, A.~Conmy, and N.~Nanda, ``Understanding
  reasoning in thinking language models via steering vectors,'' 2025. [Online].
  Available: \url{https://arxiv.org/abs/2506.18167}
\BIBentrySTDinterwordspacing

\end{thebibliography}

\appendix

\section{CounterLogic Dataset Details}
\label{app:counterlogic}

\subsection{Hierarchical Entity Triples}
\label{app:entity-triples}

The CounterLogic dataset uses the following hierarchical entity triples, where each tuple $(a, b, c)$ denotes a strict subset relationship: $a \subset b \subset c$.

\begin{center}
\begin{tabular}{lll}
\textit{siameses} & \textit{cats} & \textit{felines} \\
\textit{labradors} & \textit{dogs} & \textit{canines} \\
\textit{sedans} & \textit{cars} & \textit{vehicles} \\
\textit{humans} & \textit{animals} & \textit{mortals} \\
\textit{cruisers} & \textit{warships} & \textit{watercrafts} \\
\textit{chickadees} & \textit{birds} & \textit{winged\_animals} \\
\textit{boeings} & \textit{planes} & \textit{aircrafts} \\
\textit{pines} & \textit{evergreens} & \textit{trees} \\
\textit{anguses} & \textit{cows} & \textit{mammals} \\
\textit{daisies} & \textit{flowers} & \textit{plants} \\
\end{tabular}
\end{center}

\subsection{Sentence Templates}
\label{app:sentence-templates}

From each entity triplet, we generate sentence pairs corresponding to one of four logical sentence templates of the form $S$ and $\neg S$, capturing contradictory or complementary quantifier relations:

\begin{enumerate}
    \item \textit{All \{A\} are \{B\}}, \textit{Some \{A\} are not \{B\}}
    \item \textit{No \{A\} are \{B\}}, \textit{Some \{A\} are \{B\}}
    \item \textit{Some \{A\} are \{B\}}, \textit{No \{A\} are \{B\}}
    \item \textit{Some \{A\} are not \{B\}}, \textit{All \{A\} are \{B\}}
\end{enumerate}

\subsection{Dataset Statistics}
\label{app:dataset stats}
The final \textit{CounterLogic} dataset consists of 1,800 examples, with 200 instances for each of the 9 logical schemas listed in Table~\ref{tab:logic-forms}. To ensure a comprehensive and balanced design, four criteria were enforced during dataset construction:

\begin{itemize}
    \item \textbf{Knowledge alignment Balance}: Each logical schema contains 50\% of examples where the conclusion and the premises are knowledge-consistent under human priors and 50\% where it is not. eg.: ``All humans are animals'' is a Knowledge-Aligned statement whereas ``Some cats are not felines'' is a Knowledge-Conflicting statement.
    
    \item \textbf{Validity Balance}: Half the examples per schema are logically valid, while the remaining half intentionally violate the logical structure. eg.: ``Premise: All cats are dogs; Conclusion: Some cats are dogs'' is a valid conclusion whereas ``Premise: All cats are dogs; Conclusion: Some cats are not dogs'' is an invalid conclusion.
    
    \item \textbf{Entity Relationship Balance}: 50\% of examples involve an entity A that is a subset of entity B (e.g., \textit{siameses} $\subset$ \textit{cats}) and 50\% feature B as a subset of A (e.g., \textit{cats} $\subset$ \textit{siameses}).
    
    \item \textbf{Sentence Template Balance}: All 8 sentence-pair templates are applied evenly across examples within each logical schema, promoting lexical and syntactic diversity.
\end{itemize}

\begin{table*}[htbp]
  \centering
  \begin{tabular}{|c|p{10cm}|}
    \hline
    \textbf{Name} & \textbf{Propositional Logic Form} \\
    \hline
    Modus Ponens (MP) & $((p \to q) \land p) \vdash q$ \\
    Modus Tollens (MT) & $((p \to q) \land \neg q) \vdash \neg p$ \\
    Hypothetical Syllogism (HS) & $((p \to q) \land (q \to r)) \vdash (p \to r)$ \\
    Disjunctive Syllogism (DS) & $((p \lor q) \land \neg p) \vdash q$ \\
    Constructive Dilemma (CD) & $(p \to q) \land (r \to s) \land (p \lor r) \vdash (q \lor s)$ \\
    Destructive Dilemma (DD) & $((p \to q) \land (r \to s) \land (\neg q \lor \neg s)) \vdash (\neg p \lor \neg r)$ \\
    Bidirectional Dilemma (BD) & $((p \to q) \land (r \to s) \land (p \lor \neg s)) \vdash (q \lor \neg r)$ \\
    Commutation (CT) & $p \vdash (q \lor p)$ \\
    Material Implication (MI) & $(p \to q) \vdash (\neg p \lor q)$ \\
    \hline
  \end{tabular}
  \caption{%
    Formal propositional logical schemas used in the \textit{CounterLogic} dataset. Each row presents a canonical logical inference rule and its structure in propositional form. Believability is achieved when both the premises and the conclusion are true independently. Invalid datapoints are created by replacing conclusion statements (e.g., $p$, $q$, $r$) with unrelated ones (e.g., $p'$, $q'$, $r'$) making the logical rule invalid.
  }
  \label{tab:logic-forms}
\end{table*}

\subsection{Varying Reasoning Depth}
\label{app:varying-depth}

To systematically evaluate how logical reasoning performance scales with complexity, we implemented a method for varying the reasoning depth for certain logical schemas. For any schema that includes an implication rule of the form $p \rightarrow q$, we can increase the number of reasoning steps by replacing it with a longer chain of implications. For instance, a single implication can be expanded by introducing $i$ intermediate propositions ($x_1, x_2, \dots, x_i$) to form a chain:
$p \rightarrow x_1 \rightarrow x_2 \rightarrow \dots \rightarrow x_i \rightarrow q$.
This transformation increases the number of premises in the problem, requiring the model to perform a longer series of deductions to arrive at the final conclusion.

To preserve the original knowledge alignment of the implication, the intermediate propositions ($x_1, x_2, \dots$) introduced are always sampled from a pool of knowledge-aligned (factually true) sentences. This ensures that if a base implication $p \rightarrow q$ was knowledge-conflicting (e.g., where a true premise $p$ leads to a false conclusion $q$), at least one of the new implications in the extended chain will also be knowledge-conflicting. For example, in the chain $p \rightarrow x_1 \rightarrow q$, the link $x_1 \rightarrow q$ would be factually false, as its premise ($x_1$) is true while its conclusion ($q$) is false, thereby preserving the overall knowledge-conflicting nature of the reasoning task.

This technique is applied to all schemas in Table 2 that contain one or more implication rules. Notably, this excludes the \textit{Disjunctive Syllogism (DS)} and \textit{Commutation (CT)} schemas, as their propositional logic forms do not involve implication.

Below are two examples illustrating how base rules are modified to increase reasoning depth.

\paragraph{Example 1: Modus Ponens with One Addition}
The base Modus Ponens (MP) rule requires one step of deduction from two premises. By adding one intermediate step ($x_1$), the rule now requires two steps of deduction from three premises.
\begin{itemize}
    \item \textbf{Base MP Rule:} 
    $$((p \rightarrow q) \land p) \vdash q$$
    \item \textbf{MP Rule with One Addition:} 
    $$((p \rightarrow x_1) \land (x_1 \rightarrow q) \land p) \vdash q$$
\end{itemize}

\paragraph{Example 2: Destructive Dilemma with Two Additions}
The Destructive Dilemma (DD) rule contains two separate implication rules ($p \rightarrow q$ and $r \rightarrow s$). We add one intermediate step to each, for a total of two additions, increasing the number of premises from three to five.
\begin{itemize}
    \item \textbf{Base DD Rule:} 
    $$((p \rightarrow q) \land (r \rightarrow s) \land (\neg q \lor \neg s)) \vdash (\neg p \lor \neg r)$$
    
    \item \textbf{DD Rule with Two Additions:} 
    \begin{multline*}
        ((p \rightarrow x_1) \land (x_1 \rightarrow q) \land (r \rightarrow y_1) \\
        \land (y_1 \rightarrow s)
        \land (\neg q \lor \neg s)) \vdash (\neg p \lor \neg r)
    \end{multline*}
\end{itemize}

\subsection{Prompt for Natural Language Reformulation of Logical Structures}
\label{app:NL-conversion}

To prevent language models from relying on memorized patterns of formal logic structures, we reformulate logical premises and conclusions into natural language contexts and questions. Specifically, we prompt \textbf{GPT-4o} to carry out this transformation. The model is instructed to rewrite the given premise into a natural language context and the conclusion into a straightforward question, without preserving the surface structure of the original logical form. The transformation ensures the resulting question does not include meta-references like ``in this context,'' and the phrasing is natural and intuitive:

\begin{lstlisting}[basicstyle=\ttfamily\small, breaklines=true]

premise: [premise]
conclusion: [conclusion]
premise list: [premise list]
Make the premise into a context which is like a natural language way of writing the premises. Make conclusion into a question.
The context/questions shouldn't be too complicated but shouldn't directly be like premise/conclusion either. The question must be asked normally without stating things like "in this context" or "with this information".
premise list is only given for your better understanding.
Reply ONLY with a json with two keys 'context' and 'question'
\end{lstlisting}

\section{Prompting Strategies}
\label{app:prompting-stragies}
We evaluate three distinct prompting strategies across all tasks:

\textbf{1. Standard Condition: }
In this baseline condition, models receive direct questions with minimal guidance, instructed to consider only the logical validity of arguments regardless of premise believability:

\begin{lstlisting}[basicstyle=\ttfamily\small, breaklines=true]

  Based on the following premises, determine if the conclusion logically follows. Consider only the logical validity based on the given premises, regardless of whether the premises themselves are factually true.

  Premises:
    1. [Premise 1]
    2. [Premise 2]
    
    Conclusion: [Conclusion]
    
    Does the conclusion logically follow from the premises? Answer with "Yes" or "No" and explain your reasoning step by step.
\end{lstlisting}

\begin{table*}[t]
\centering
\begin{tabularx}{\linewidth}{|l|X|X|X|}
\hline
\textbf{Task} & \textbf{Dataset Content} & \textbf{Initial Reflection Input} & \textbf{Reasoning Input} \\ 
\hline
Syllogisms & Two premises + conclusion & Conclusion statement & Full syllogism \\
\hline
KNOT & Passage + Q/A pair & Answer without passage & Full passage + Q/A \\
\hline
FOLIO & Narrative + claim & Isolated claim & Complete narrative \\
\hline
LogicBench & Context + Q/A & Q/A without context & Full context + Q/A \\
\hline
Arithmetic & Base equation & Equation without base & Base-specified equation \\
\hline
\end{tabularx}
\caption{\textbf{Task-Specific Implementation of Two-Stage Reflection Approach.} This table outlines how our reflective prompting strategy is applied across different task types. \textbf{Initial Reflection Input} refers to the isolated information presented for knowledge alignment assessment. \textbf{Reasoning Input} shows the complete information provided in the second stage for logical assessment. This separation helps models distinguish between plausibility assessment and formal logical analysis.}

\label{tab:task-approaches}
\end{table*}

\textbf{2. Metacognitive Condition: }
In this condition, we introduce a preliminary reflection step (asking the model what it thinks about a statement) before the reasoning task. 

\begin{lstlisting}[basicstyle=\ttfamily\small, breaklines=true]

Prompt 1:

Is the following statement factually correct?

statement: [Conclusion]

Answer only with Yes or No.

Prompt 2(in the same context):

Now, based on the  following premises, determine if the conclusion logically follows. Consider only the logical validity based on the
given premises, regardless of whether the premises themselves are factually true.
Premises:
    1. [Premise 1]
    2. [Premise 2]
    
    Conclusion: [Conclusion]
    
    Does the conclusion logically follow from the premises? Answer with "Yes" or "No" and explain your reasoning step by step.
\end{lstlisting}

The above is the general structure of our prompting method. The prompts are modified according to the dataset we evaluate.



\section{Logical Inference Schemas}
\label{app:logical-forms}

The \textit{CounterLogic} dataset uses various formal propositional logic inference schemas to generate reasoning examples, as detailed in Table~\ref{tab:logic-forms}.

\section{Task-Specific Reflection Approaches}
\label{app:task-approaches}

Our metacognitive intervention is implemented with task-specific adaptations to ensure appropriate reflection across different reasoning formats.

\textbf{Hierarchical Syllogisms: }
Derived from classical syllogistic reasoning and adapted from \cite{bertolazzi2024systematicanalysislargelanguage}'s work, this task presents logically structured arguments where the conclusion may conflict with world knowledge. Each example contains two premises and a conclusion, with models evaluating logical validity. For \textit{FaR}, models first assess the conclusion statement in isolation for its alignment with parametric knowledge, then evaluate the full syllogism's logical validity(see Figure~\ref{fig:crownjewel}B).

\textbf{KNOT: }
Adapted from the Knowledge Conflict Resolution benchmark \cite{liu_untangle_2024}, this task evaluates reasoning through explicit (KNOT-E) and implicit (KNOT-I) conflict resolution. Each instance contains a passage with counterfactual information, a question, and an answer. The \textit{FaR} implementation first presents the answer in isolation for plausibility assessment, then provides the full passage and question-answer pair for contextual reasoning. This separation tests models' ability to distinguish between prior knowledge and contextual truth.

\textbf{FOLIO: }
Using long-form deductive reasoning problems from FOLIO \cite{han2024folionaturallanguagereasoning}, this task requires evaluating whether conclusions logically follow from multi-step narratives. Our \textit{FaR} approach first presents the conclusion for isolated plausibility judgment, then provides the complete narrative for logical analysis. 

\textbf{LogicBench: }
This reasoning dataset \cite{parmar2024logicbench} combines first-order, non-monotonic, and propositional logic problems. It tests models' ability to follow formal logical rules while overriding potentially conflicting parametric knowledge. The \textit{FaR} implementation presents questions and answers without supporting context for initial plausibility assessment, followed by complete logical contexts for formal evaluation.


\textbf{CounterLogic: }
For our novel benchmark, described in Section \ref{sec:counterlogic}, we apply the same two-stage reflection approach used in the Hierarchical Syllogisms task, first assessing conclusion plausibility in isolation before evaluating logical validity within the full syllogistic context.

Knowledge alignment labels were available only for the \textit{CounterLogic} and \textit{Hierarchical Syllogisms} datasets. For the remaining datasets, we get alignment judgments by prompting the LLMs to assess the factuality of the conclusions, and we use their responses as proxy labels. This procedure results in a heavily imbalanced label distributions, most notably in the \textit{KNOT} dataset, where the ratio of Aligned to Conflicting cases is approximately 1:15.
\label{app:ground-truth-alignment}

\section{Models}
\label{app:models}

In our study, we evaluated 11 state-of-the-art large language models from various organizations, spanning different architectures, parameter scales, and training paradigms. Below we provide details about each model, including their version, size, and key characteristics:

\begin{table*}[ht]
\centering
\caption{Details of the evaluated models}
\label{tab:model-details}
\begin{tabularx}{\textwidth}{lXll}
\toprule
\textbf{Model} & \textbf{Developer} & \textbf{Parameters} & \textbf{Release Date} \\
\midrule
GPT-4o & OpenAI & Unknown & May 2024 \\
GPT-4o‑mini & OpenAI & Unknown & July 2024 \\
o3-mini & OpenAI & Unknown & April 2025 \\
Gemini‑Flash‑1.5 & Google DeepMind & Unknown & May 2024 \\
Llama‑3.3‑70B & Meta AI & 70B & December 2024 \\
Llama‑3.1‑70B & Meta AI & 70B & July 2024 \\
Llama‑3.1‑8B & Meta AI & 8B & July 2024 \\
Qwen‑2.5‑72B & Alibaba & 72B & September 2024 \\
Qwen‑2.5‑7B & Alibaba & 7B & September 2024 \\
DeepSeek‑V3 & DeepSeek AI & 671B & Jan 2025 \\
Deepseek‑R1‑distill‑Llama & DeepSeek AI & 671B & January 2025 \\
\bottomrule
\end{tabularx}
\end{table*}

\subsection{Model Access}
All models were accessed through the OpenRouter API to ensure consistent evaluation conditions. This approach allowed us to standardize the inference parameters across different model providers, including temperature settings (0.7), top-p (0.95), and maximum token length (4096 tokens).

\subsection{Model Selection Criteria}
We selected these models based on the following criteria:

\begin{enumerate}
    \item \textbf{State-of-the-art performance:} All selected models represent the cutting edge of LLM development at the time of our study.
    
    \item \textbf{Architectural diversity:} We included models with different architectural designs to examine whether the observed patterns generalize across various model architectures.
    
    \item \textbf{Parameter scale variation:} The selection spans from relatively smaller models (7B parameters) to much larger ones (72B+ parameters) to investigate how model size correlates with counterfactual reasoning abilities.
    
    \item \textbf{Training paradigm diversity:} The models employ various training approaches, including different pretraining datasets, fine-tuning strategies, and alignment techniques.
\end{enumerate}

\subsection{Model Specifications}

\subsubsection{OpenAI Models}
\textbf{GPT-4o}, \textbf{GPT-4o-mini} and \textbf{o3-mini} represents the OpenAI's multimodal model designed for both text and imageprocessing. GPT-4o is the standard non reasoning model provided by OpenAI. GPT-4o mini is a lightweight and faster version of 4o. o3 mini is the reasoning model provided by openAI that uses specialized token to internally do CoT before answering.

\subsubsection{Google Models}
\textbf{Gemini-Flash-1.5} is the lightweight version of Google's Gemini 1.5 model family, optimized for quick responses while maintaining strong reasoning capabilities.

\subsubsection{Meta AI Models}
\textbf{Llama-3.3-70B}, \textbf{Llama-3.1-70B}, and \textbf{Llama-3.1-8B} represent Meta AI's open-source LLM efforts. The 3.1 series is an upgrade to the Llama-3 models, with enhanced instruction-following and reasoning capabilities. We include both larger (70B) and smaller (8B) parameter variants to examine scaling effects.

\subsubsection{Alibaba Models}
\textbf{Qwen-2.5-72B} and \textbf{Qwen-2.5-7B} are Alibaba's latest generation language models, known for their strong performance across various benchmarks, particularly in multilingual reasoning tasks.

\subsubsection{DeepSeek Models}
\textbf{DeepSeek-V3} is a 671B parameter model developed by DeepSeek AI, designed specifically for dialogue applications with strong reasoning capabilities.

\textbf{DeepSeek-R1-Distill-Llama} is reasoning model that is finetuned version of DeepSeek-R1(671b) using the outputs of Llama-3.3-70b-instruct model.

\subsection{Model Inference Parameters}
For all evaluations, we used consistent inference parameters across models. We use openRouter for all the non OpenAI models and OpenAI API for the  3 openAI models. All the models that support system prompts have standard prompt asking for instruction following.

\subsection{Cost of the Evaluations}
Overall 83.2\$ were spend on openRouter for all non-OpenAI models across all tasks.\\
About 2000\$ of OpenAI credits were used for running the evaluation, most of which was used by the o1-preview model.

\begin{figure*}
  \centering
  \includegraphics[width=\textwidth]{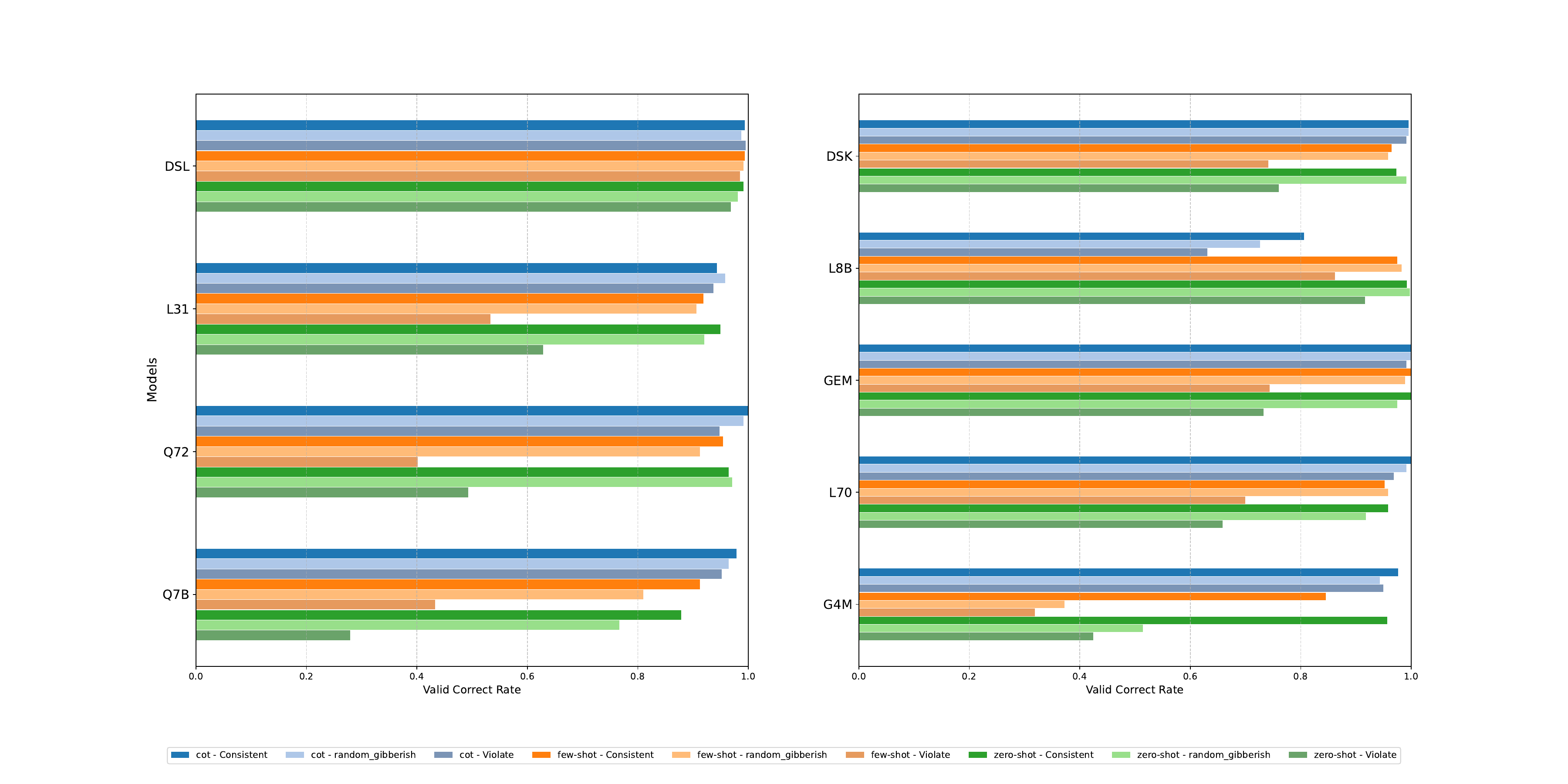}
  \caption{\textbf{Ablation study comparing the various prompting strategies.} Shows that CoT always outperforms zero-shot but few-shot fails to do so in some models for the Syllogistic dataset.}
  \label{fig:ablation-zero}
\end{figure*}

\begin{figure}[ht]
    \centering
    \includegraphics[width=\linewidth]{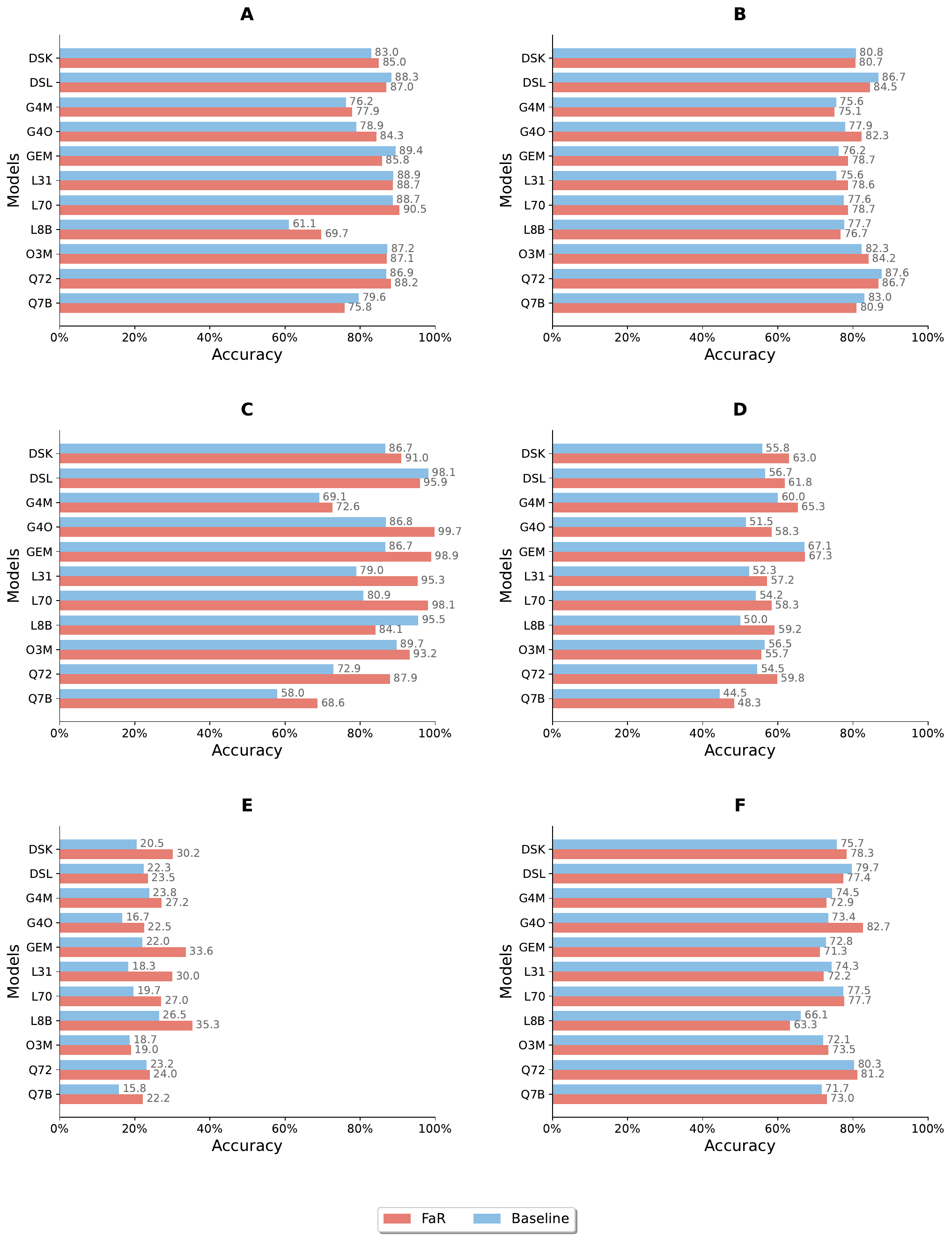} 
    \captionsetup{width=\linewidth}
    \caption{%
    \textbf{
    Accuracy comparision between the baseline setup and our \textit{FaR} setup averaged across all the datasets.} The x-axis represents accuracy and the y-axis represents the models.  We can see that our method \textit{FaR} improves performance across all the tasks.
    The y-axis represents the following models - DSK: DeepSeek Chat, DSL: Deepseek R1 Distill Llama 70B, G4M: GPT-4o Mini, G4O: GPT-4o, GEM: Google Gemini Flash 1.5, L8B: Llama 3.1 8B Instruct, L31: Llama 3.1 70B Instruct, L70: Llama 3.3 70B Instruct, O3M: o3-mini, Q7B: Qwen 2.5 7B Instruct, and Q72: Qwen 2.5 72B Instruct.
    Each of the above graphs represents the performance on the following datasets - A: CounterLogic, B: FOLIO, C: Hierarchical Syllogisms, D: KNOT-Explicit, E: KNOT-Implicit, F: Logic Bench
    } 
    \label{fig:selfseg-dataset-comparison}
\end{figure}

\begin{figure}[ht]
    \centering
    \includegraphics[width=\linewidth]{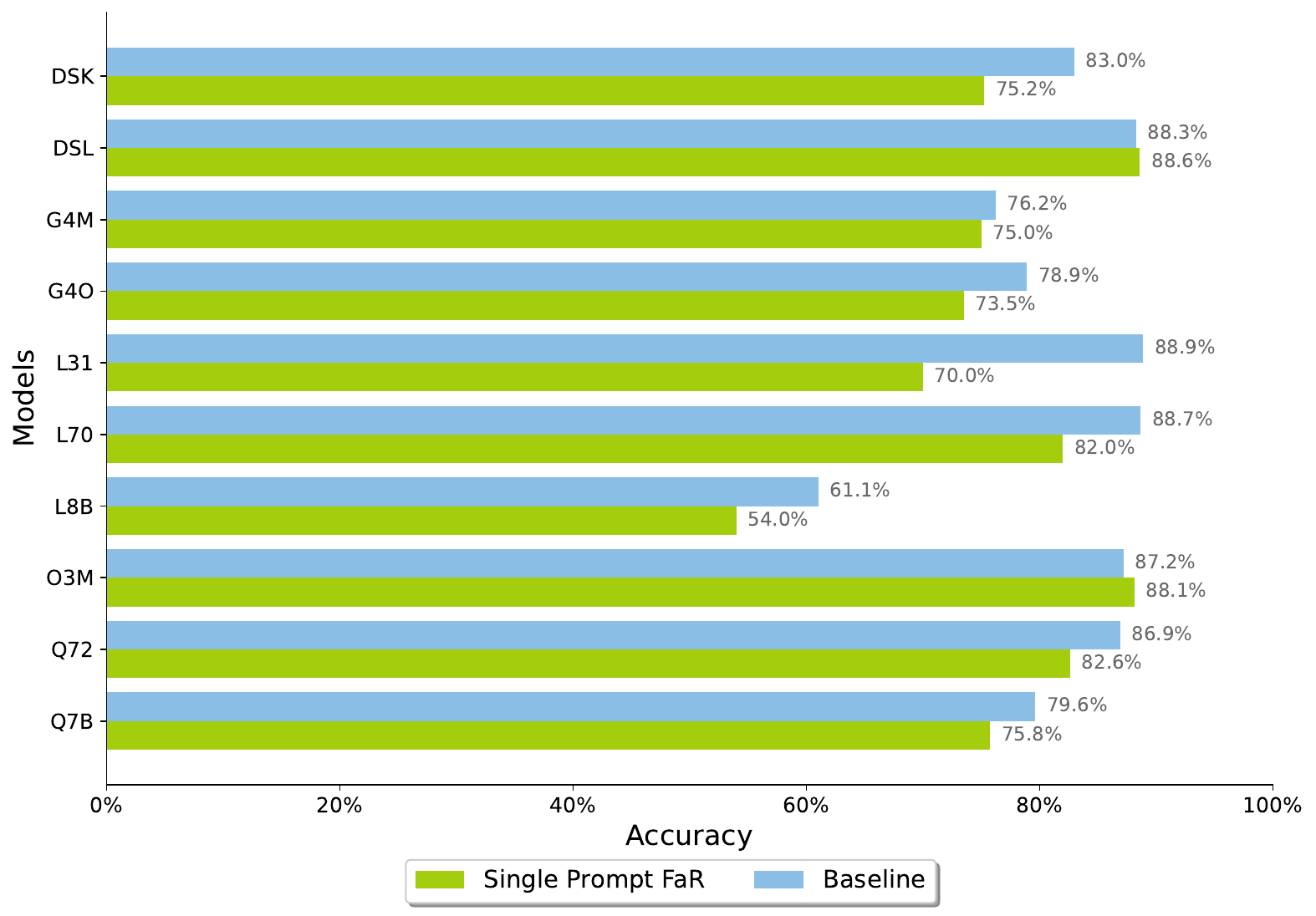} 
    \captionsetup{width=\linewidth}
    \caption{%
    \textbf{
    Accuracy comparison between the baseline setup and the Single prompt \textit{FaR} setup averaged across all the datasets.} The x-axis represents accuracy and the y-axis represents the models.  We can see that this modification of \textit{FaR} does not lead to any significant performance gains.
    } 
    \label{fig:selfseg-weird-accuracy-comparison}
\end{figure}

\section{Full Results}
\label{app:full_results}
This section provides the full experimental results for all models and evaluation settings. Table \ref{tab:model_performance} reports accuracies with and without \textit{FaR}, while Tables \ref{tab:model_performance_valid_base} and \ref{tab:model-performance-valid-far} present a comparison between Knowledge Aligned and Knowledge Conflicting accuracies on the baseline and \textit{FaR} subsets, respectively. Each value denotes the mean $\pm$ standard deviation. These tables serve as a comprehensive reference supporting the main findings discussed in paper.

\section{Model-wise Results and Performance Patterns}

Our evaluation demonstrates that the \textit{FaR} method provides a consistent and significant improvement in reasoning accuracy across all 11 evaluated models when averaged across the datasets. This universal performance boost is clearly illustrated in Figure 6, where the FaR method (pink bar) outperforms the baseline (blue bar) for every model tested. The most noticeable average improvement was observed for the GPT-4o (G4O) model, which saw an accuracy increase of approximately 7\%.

\subsection{Key Performance Patterns}
A deeper analysis of the results, particularly from the detailed breakdown in Table 5, reveals several important patterns in how different models respond to the FaR intervention.

\begin{itemize}
    \item \textbf{Variable Impact Across Datasets:} While the FaR method is universally beneficial, its impact varies significantly depending on the reasoning task. 
        \begin{itemize}
            \item The most dramatic performance gains are seen on the \textbf{Hierarchical} syllogisms dataset, where accuracy for many models, such as L70 (80.9\% to 98.1\%) and G4O (86.8\% to 99.7\%), improved by over 15 percentage points.
            \item In contrast, the improvement on the \textbf{Folio} datasets is more modest, with FaR providing a smaller, though still consistent, accuracy boost emphasizing the need for better conflict resolution strategies for complex narrative tasks
        \end{itemize}

    \item \textbf{Effectiveness Across Model Types:} The benefits of FaR are not limited to a specific class or size of model.
        \begin{itemize}
            \item Specialized \textbf{Reasoning Models} like O3M and DSL, which already have strong baseline performance, still achieve slightly higher accuracy with the FaR intervention. For example, O3M's performance on the Hierarchical task improves from 89.7\% to 93.2\%.
            \item Similarly, \textbf{Small Models} such as L8B and Q7B show significant gains, indicating that the method is a broadly applicable technique for enhancing logical robustness, regardless of model scale.
        \end{itemize}
\end{itemize}

In summary, the FaR intervention proves to be a robust and widely effective method for improving logical reasoning. The degree of improvement, however, is task-dependent, highlighting its particular strength in mitigating the types of errors prevalent in tasks with strong knowledge conflicts.

\begin{table*}[t]
\centering
\resizebox{\textwidth}{!}{
\begin{tabular}{l@{\hspace{1em}}cc|cc|cc|cc|cc|cc}
\toprule
& \multicolumn{2}{c|}{Explicit} & \multicolumn{2}{c|}{Implicit} & \multicolumn{2}{c|}{Folio} & \multicolumn{2}{c|}{Hierarchical} & \multicolumn{2}{c|}{Counterlogic} & \multicolumn{2}{c|}{Logic Bench} \\
\cmidrule(lr){2-3} \cmidrule(lr){4-5} \cmidrule(lr){6-7} \cmidrule(lr){8-9} \cmidrule(lr){10-11} \cmidrule(lr){12-13}
\textbf{Models} & FaR & Baseline & FaR & Baseline & FaR & Baseline & FaR & Baseline & FaR & Baseline & FaR & Baseline \\
\midrule
\multicolumn{13}{l}{\textit{Large models}} \\
L70 & $\textbf{58.3} \pm 0.7$ & $54.2 \pm 0.2$ & $\textbf{27.0} \pm 0.7$ & $19.7 \pm 1.3$ & $\textbf{78.7} \pm 2.2$ & $77.6 \pm 1.1$ & $\textbf{98.1} \pm 0.1$ & $80.9 \pm 0.1$ & $\textbf{90.5} \pm 0.2$ & $88.7 \pm 0.1$ & $\textbf{77.7} \pm 0.4$ & $77.5 \pm 0.2$ \\
L31 & $\textbf{57.2} \pm 0.2$ & $52.3 \pm 0.3$ & $\textbf{30.0} \pm 1.0$ & $18.3 \pm 0.3$ & $\textbf{78.6} \pm 0.1$ & $75.6 \pm 2.0$ & $\textbf{95.3} \pm 0.5$ & $79.0 \pm 0.6$ & $88.7 \pm 0.2$ & $\textbf{88.9} \pm 0.4$ & $72.2 \pm 0.6$ & $\textbf{74.3} \pm 0.3$ \\
GEM & $\textbf{67.3} \pm 0.3$ & $67.1 \pm 0.3$ & $\textbf{33.6} \pm 0.6$ & $22.0 \pm 0.0$ & $\textbf{78.7} \pm 0.6$ & $76.2 \pm 0.6$ & $\textbf{98.9} \pm 0.0$ & $86.7 \pm 0.0$ & $85.8 \pm 0.2$ & $\textbf{89.4} \pm 0.2$ & $71.3 \pm 0.2$ & $\textbf{72.8} \pm 0.3$ \\
Q72 & $\textbf{59.8} \pm 0.5$ & $54.5 \pm 0.8$ & $\textbf{24.0} \pm 0.3$ & $23.2 \pm 0.2$ & $86.7 \pm 0.4$ & $\textbf{87.6} \pm 1.2$ & $\textbf{87.9} \pm 0.2$ & $72.9 \pm 0.5$ & $\textbf{88.2} \pm 0.8$ & $86.9 \pm 0.4$ & $\textbf{81.2} \pm 0.5$ & $80.3 \pm 0.1$ \\
DSK & $\textbf{63.0} \pm 1.0$ & $55.8 \pm 0.2$ & $\textbf{30.2} \pm 0.5$ & $20.5 \pm 1.8$ & $80.7 \pm 1.1$ & $\textbf{80.8} \pm 1.0$ & $\textbf{91.0} \pm 0.8$ & $86.7 \pm 0.6$ & $\textbf{85.0} \pm 0.6$ & $83.0 \pm 0.4$ & $\textbf{78.3} \pm 0.0$ & $75.7 \pm 0.0$ \\
G4O & $\textbf{58.3} \pm 0.3$ & $51.5 \pm 0.2$ & $\textbf{22.5} \pm 0.2$ & $16.7 \pm 0.0$ & $\textbf{80.8} \pm 3.9$ & $78.1 \pm 1.7$ & $\textbf{99.7} \pm 0.1$ & $86.8 \pm 0.1$ & $\textbf{84.3} \pm 0.9$ & $78.9 \pm 0.7$ & $\textbf{82.7} \pm 0.3$ & $73.4 \pm 0.4$ \\
\midrule
\multicolumn{13}{l}{\textit{Small models}} \\
L8B & $\textbf{59.2} \pm 1.8$ & $50.0 \pm 2.3$ & $\textbf{35.3} \pm 2.7$ & $26.5 \pm 1.8$ & $76.7 \pm 0.5$ & $\textbf{77.7} \pm 2.1$ & $84.1 \pm 0.7$ & $\textbf{95.5} \pm 1.2$ & $\textbf{69.7} \pm 1.2$ & $61.1 \pm 1.7$ & $63.3 \pm 0.7$ & $\textbf{66.2} \pm 0.6$ \\
Q7B & $\textbf{48.3} \pm 0.3$ & $44.5 \pm 1.2$ & $\textbf{22.2} \pm 0.8$ & $15.8 \pm 0.8$ & $80.9 \pm 2.3$ & $\textbf{83.0} \pm 1.3$ & $\textbf{68.6} \pm 0.1$ & $58.0 \pm 0.4$ & $75.8 \pm 0.2$ & $\textbf{79.6} \pm 0.4$ & $\textbf{73.0} \pm 0.3$ & $71.7 \pm 1.0$ \\
G4M & $\textbf{65.3} \pm 2.0$ & $60.0 \pm 0.3$ & $\textbf{27.2} \pm 0.5$ & $23.8 \pm 0.2$ & $75.1 \pm 1.6$ & $\textbf{75.1} \pm 1.5$ & $\textbf{72.6} \pm 0.2$ & $69.1 \pm 0.7$ & $\textbf{77.9} \pm 0.4$ & $76.2 \pm 0.3$ & $72.9 \pm 0.4$ & $\textbf{74.4} \pm 0.4$ \\
\midrule
\multicolumn{13}{l}{\textit{Reasoning models}} \\
DSL & $\textbf{61.8} \pm 0.2$ & $56.7 \pm 0.3$ & $\textbf{23.5} \pm 0.5$ & $22.3 \pm 0.7$ & $84.5 \pm 1.5$ & $\textbf{86.7} \pm 0.2$ & $95.9 \pm 1.2$ & $\textbf{98.1} \pm 0.1$ & $87.0 \pm 0.3$ & $\textbf{88.3} \pm 0.8$ & $77.4 \pm 0.0$ & $\textbf{79.7} \pm 0.7$ \\
O3M & $55.7 \pm 0.0$ & $\textbf{56.5} \pm 0.2$ & $\textbf{19.0} \pm 0.3$ & $18.7 \pm 0.3$ & $\textbf{84.2} \pm 0.0$ & $82.3 \pm 0.1$ & $\textbf{93.2} \pm 0.6$ & $89.7 \pm 0.5$ & $86.9 \pm 0.4$ & $\textbf{87.2} \pm 0.4$ & $\textbf{73.5} \pm 0.3$ & $72.1 \pm 1.1$ \\
\bottomrule
\end{tabular}
}
\caption{Model performance (FaR vs. Baseline) across datasets, showing mean accuracy and standard deviation (\%)}
\label{tab:model_performance}

\end{table*}

\begin{table*}[t]
\centering

\resizebox{\textwidth}{!}{
\begin{tabular}{l@{\hspace{1em}}cc|cc|cc|cc|cc|cc}
\toprule
& \multicolumn{2}{c|}{Implicit} & \multicolumn{2}{c|}{Explicit} & \multicolumn{2}{c|}{Folio} & \multicolumn{2}{c|}{Hierarchical} & \multicolumn{2}{c|}{Counterlogic} & \multicolumn{2}{c|}{Logic Bench} \\
\cmidrule(lr){2-3} \cmidrule(lr){4-5} \cmidrule(lr){6-7} \cmidrule(lr){8-9} \cmidrule(lr){10-11} \cmidrule(lr){12-13}
\textbf{Models} & Aligned & Conflicting & Aligned & Conflicting & Aligned & Conflicting & Aligned & Conflicting & Aligned & Conflicting & Aligned & Conflicting \\
\midrule
\multicolumn{13}{l}{\textit{Large models}} \\
L70 & $\textbf{25.4} \pm 2.9$ & $18.8 \pm 1.1$ & $47.0 \pm 0.4$ & $\textbf{55.1} \pm 0.3$ & $\textbf{91.4} \pm 1.0$ & $64.3 \pm 1.7$ & $\textbf{95.8} \pm 0.6$ & $65.9 \pm 0.4$ & $83.2 \pm 0.5$ & $\textbf{84.0} \pm 0.4$ & $\textbf{78.4} \pm 1.2$ & $63.8 \pm 0.3$ \\
L31 & $\textbf{24.3} \pm 3.8$ & $17.6 \pm 0.0$ & $43.2 \pm 3.2$ & $\textbf{53.2} \pm 0.1$ & $\textbf{86.0} \pm 0.5$ & $67.2 \pm 3.3$ & $\textbf{95.0} \pm 0.7$ & $62.9 \pm 0.6$ & $\textbf{91.2} \pm 0.2$ & $88.7 \pm 1.0$ & $\textbf{75.2} \pm 1.3$ & $59.1 \pm 0.1$ \\
GEM & $21.3 \pm 1.3$ & $\textbf{22.2} \pm 0.2$ & $\textbf{68.3} \pm 0.0$ & $67.0 \pm 0.3$ & $\textbf{85.7} \pm 1.7$ & $66.3 \pm 0.5$ & $\textbf{100.0} \pm 0.0$ & $73.3 \pm 0.0$ & $79.0 \pm 0.4$ & $\textbf{82.5} \pm 0.4$ & $\textbf{81.9} \pm 1.3$ & $58.9 \pm 0.5$ \\
Q72 & $\textbf{44.4} \pm 2.5$ & $20.7 \pm 0.1$ & $\textbf{68.3} \pm 6.7$ & $53.9 \pm 0.6$ & $\textbf{93.0} \pm 1.2$ & $83.8 \pm 1.9$ & $\textbf{96.5} \pm 0.3$ & $49.3 \pm 1.0$ & $\textbf{83.5} \pm 0.7$ & $77.3 \pm 2.3$ & $\textbf{86.4} \pm 1.8$ & $68.0 \pm 0.1$ \\
DSK & $\textbf{36.8} \pm 0.8$ & $19.1 \pm 2.0$ & $\textbf{60.0} \pm 5.0$ & $55.5 \pm 0.5$ & $\textbf{90.4} \pm 1.3$ & $75.0 \pm 1.4$ & $\textbf{97.4} \pm 0.2$ & $76.0 \pm 1.2$ & $\textbf{74.7} \pm 1.5$ & $65.2 \pm 1.0$ & $\textbf{91.0} \pm 0.7$ & $65.0 \pm 0.9$ \\
G4O & $\textbf{34.9} \pm 0.6$ & $14.5 \pm 0.0$ & $\textbf{59.1} \pm 2.0$ & $51.0 \pm 0.3$ & $\textbf{88.1} \pm 2.7$ & $68.1 \pm 2.5$ & $\textbf{97.4} \pm 0.3$ & $76.1 \pm 0.1$ & $\textbf{64.8} \pm 0.8$ & $58.6 \pm 1.6$ & $\textbf{86.4} \pm 2.6$ & $70.7 \pm 0.3$ \\
\midrule
\multicolumn{13}{l}{\textit{Small models}} \\
L8B & $\textbf{27.9} \pm 0.8$ & $24.9 \pm 4.6$ & $49.6 \pm 1.4$ & $\textbf{50.3} \pm 3.4$ & $\textbf{80.8} \pm 4.3$ & $75.5 \pm 1.3$ & $\textbf{99.2} \pm 0.2$ & $91.7 \pm 2.5$ & $\textbf{55.3} \pm 1.2$ & $52.2 \pm 1.5$ & $\textbf{62.5} \pm 0.0$ & $57.9 \pm 2.3$ \\
Q7B & $\textbf{17.8} \pm 2.2$ & $15.4 \pm 1.4$ & $\textbf{82.1} \pm 10.7$ & $42.3 \pm 1.4$ & $\textbf{86.6} \pm 2.0$ & $81.8 \pm 1.1$ & $\textbf{87.9} \pm 0.5$ & $28.0 \pm 0.4$ & $\textbf{64.2} \pm 1.2$ & $56.8 \pm 0.2$ & $\textbf{83.9} \pm 6.5$ & $51.6 \pm 1.3$ \\
G4M & $\textbf{45.1} \pm 3.6$ & $20.6 \pm 0.3$ & $\textbf{66.7} \pm 4.7$ & $59.3 \pm 0.8$ & $\textbf{85.5} \pm 1.0$ & $66.4 \pm 2.5$ & $\textbf{95.7} \pm 0.3$ & $42.5 \pm 1.2$ & $\textbf{66.3} \pm 0.7$ & $55.6 \pm 1.0$ & $\textbf{80.8} \pm 1.3$ & $59.0 \pm 1.3$ \\
\midrule
\multicolumn{13}{l}{\textit{Reasoning models}} \\
DSL & $\textbf{42.7} \pm 2.7$ & $19.9 \pm 1.2$ & $43.0 \pm 9.6$ & $\textbf{57.4} \pm 0.1$ & $\textbf{94.6} \pm 0.0$ & $79.3 \pm 0.6$ & $\textbf{98.6} \pm 0.5$ & $97.6 \pm 0.7$ & $\textbf{86.6} \pm 0.6$ & $82.7 \pm 2.0$ & $\textbf{85.9} \pm 1.3$ & $77.2 \pm 1.4$ \\
O3M & $\textbf{35.2} \pm 0.8$ & $16.9 \pm 0.5$ & $\textbf{60.5} \pm 2.6$ & $56.2 \pm 0.4$ & $\textbf{90.9} \pm 0.5$ & $76.1 \pm 0.5$ & $89.7 \pm 0.1$ & $89.7 \pm 0.9$ & $\textbf{83.3} \pm 0.6$ & $81.2 \pm 1.0$ & $\textbf{84.3} \pm 0.8$ & $69.0 \pm 1.2$ \\
\bottomrule
\end{tabular}
}

\caption{Model performance (Aligned vs. Conflicting) on valid datapoints with standard CoT method (baseline).}
\label{tab:model_performance_valid_base}
\end{table*}

\begin{table*}[t]
\centering
\resizebox{\textwidth}{!}{
\begin{tabular}{l@{\hspace{1em}}cc|cc|cc|cc|cc|cc}
\toprule
& \multicolumn{2}{c|}{Implicit} & \multicolumn{2}{c|}{Explicit} & \multicolumn{2}{c|}{Folio} & \multicolumn{2}{c|}{Hierarchical} & \multicolumn{2}{c|}{Counterlogic} & \multicolumn{2}{c|}{Logic Bench} \\
\cmidrule(lr){2-3} \cmidrule(lr){4-5} \cmidrule(lr){6-7} \cmidrule(lr){8-9} \cmidrule(lr){10-11} \cmidrule(lr){12-13}
\textbf{Models} & Aligned & Conflicting & Aligned & Conflicting & Aligned & Conflicting & Aligned & Conflicting & Aligned & Conflicting & Aligned & Conflicting \\
\midrule
\multicolumn{13}{l}{\textit{Large models}} \\
L70 & $\textbf{31.7} \pm 0.8$ & $26.3 \pm 0.7$ & $51.3 \pm 1.3$ & $\textbf{59.2} \pm 0.8$ & $\textbf{90.6} \pm 1.0$ & $67.5 \pm 4.6$ & $\textbf{98.4} \pm 0.2$ & $97.5 \pm 0.1$ & $87.7 \pm 0.9$ & $\textbf{92.0} \pm 0.8$ & $81.5 \pm 0.5$ & $\textbf{84.3} \pm 0.2$ \\
L31 & $\textbf{34.7} \pm 1.4$ & $29.4 \pm 1.3$ & $38.9 \pm 3.2$ & $\textbf{58.8} \pm 0.8$ & $\textbf{87.4} \pm 2.8$ & $72.1 \pm 2.2$ & $\textbf{95.4} \pm 0.3$ & $95.0 \pm 0.9$ & $92.2 \pm 0.6$ & $\textbf{94.3} \pm 0.2$ & $70.6 \pm 2.8$ & $\textbf{84.5} \pm 0.9$ \\
GEM & $\textbf{58.0} \pm 1.4$ & $29.7 \pm 0.5$ & $\textbf{72.6} \pm 2.5$ & $66.9 \pm 0.5$ & $\textbf{85.1} \pm 0.7$ & $72.0 \pm 0.9$ & $\textbf{99.6} \pm 0.0$ & $98.0 \pm 0.1$ & $\textbf{81.2} \pm 0.5$ & $80.3 \pm 0.6$ & $\textbf{78.0} \pm 3.2$ & $62.2 \pm 0.6$ \\
Q72 & $\textbf{37.5} \pm 0.0$ & $22.4 \pm 0.4$ & $\textbf{75.7} \pm 4.3$ & $59.0 \pm 0.3$ & $\textbf{93.8} \pm 1.2$ & $81.5 \pm 1.3$ & $\textbf{94.2} \pm 0.5$ & $80.7 \pm 0.3$ & $\textbf{89.2} \pm 0.2$ & $86.8 \pm 0.8$ & $\textbf{88.3} \pm 2.0$ & $81.8 \pm 0.6$ \\
DSK & $\textbf{38.0} \pm 12.0$ & $29.4 \pm 0.6$ & $58.7 \pm 0.4$ & $\textbf{63.4} \pm 1.0$ & $\textbf{93.5} \pm 0.7$ & $72.7 \pm 1.8$ & $\textbf{95.0} \pm 0.4$ & $86.2 \pm 1.2$ & $\textbf{82.7} \pm 0.2$ & $74.8 \pm 1.0$ & $\textbf{82.8} \pm 1.8$ & $75.5 \pm 0.5$ \\
G4O & $\textbf{39.3} \pm 2.6$ & $20.6 \pm 0.5$ & $\textbf{60.7} \pm 7.7$ & $58.2 \pm 0.1$ & $\textbf{89.9} \pm 3.8$ & $72.1 \pm 4.4$ & $99.7 \pm 0.1$ & $\textbf{99.8} \pm 0.0$ & $\textbf{76.2} \pm 1.9$ & $73.5 \pm 4.2$ & $\textbf{83.0} \pm 0.9$ & $82.6 \pm 0.2$ \\
\midrule
\multicolumn{13}{l}{\textit{Small models}} \\
L8B & $30.3 \pm 2.4$ & $\textbf{40.8} \pm 3.2$ & $58.1 \pm 0.6$ & $\textbf{60.5} \pm 3.4$ & $\textbf{79.1} \pm 0.8$ & $75.3 \pm 1.2$ & $\textbf{91.8} \pm 1.5$ & $76.1 \pm 2.5$ & $66.5 \pm 1.7$ & $\textbf{67.2} \pm 0.3$ & $42.7 \pm 1.7$ & $\textbf{87.6} \pm 1.8$ \\
Q7B & $\textbf{26.5} \pm 0.5$ & $21.3 \pm 0.9$ & $\textbf{82.8} \pm 0.5$ & $46.2 \pm 0.4$ & $\textbf{90.0} \pm 3.0$ & $77.8 \pm 2.1$ & $\textbf{90.2} \pm 0.3$ & $45.6 \pm 0.2$ & $\textbf{72.8} \pm 0.8$ & $63.2 \pm 0.6$ & $\textbf{84.6} \pm 2.9$ & $59.3 \pm 0.7$ \\
G4M & $\textbf{49.2} \pm 4.5$ & $23.8 \pm 1.4$ & $\textbf{73.2} \pm 1.8$ & $64.7 \pm 2.4$ & $\textbf{87.1} \pm 1.6$ & $65.4 \pm 1.6$ & $\textbf{87.5} \pm 0.8$ & $54.6 \pm 1.3$ & $\textbf{77.2} \pm 1.2$ & $58.0 \pm 0.7$ & $76.6 \pm 1.7$ & $\textbf{85.6} \pm 1.0$ \\
\midrule
\multicolumn{13}{l}{\textit{Reasoning models}} \\
DSL & $\textbf{38.7} \pm 5.4$ & $21.6 \pm 0.1$ & $\textbf{66.6} \pm 1.9$ & $61.5 \pm 0.0$ & $\textbf{93.0} \pm 0.5$ & $76.4 \pm 2.4$ & $\textbf{96.0} \pm 1.1$ & $95.8 \pm 1.3$ & $80.0 \pm 0.8$ & $\textbf{80.7} \pm 1.3$ & $\textbf{83.3} \pm 0.1$ & $75.1 \pm 0.1$ \\
O3M & $\textbf{30.3} \pm 3.0$ & $17.8 \pm 0.6$ & $\textbf{67.8} \pm 5.9$ & $54.8 \pm 0.4$ & $\textbf{91.7} \pm 0.3$ & $78.4 \pm 0.1$ & $\textbf{93.8} \pm 0.1$ & $92.4 \pm 1.4$ & $\textbf{83.7} \pm 1.2$ & $80.3 \pm 1.6$ & $\textbf{85.1} \pm 1.8$ & $70.7 \pm 0.1$ \\
\bottomrule
\end{tabular}
}
\caption{Model performance (Aligned vs. Conflicting) on valid datapoints with standard \textit{FaR} method.}
\label{tab:model-performance-valid-far}
\end{table*}

\section{Ablation Studies}
\label{app:ablations}

\subsection{Comparison of Prompting Strategies}

Figure \ref{fig:ablation-zero} presents an ablation study evaluating model performance under three prompting strategies: zero-shot, few-shot, and chain-of-thought (CoT) across three belief consistency conditions; aligned, conflicting, and random gibberish. Here random gibberish datapoints are obtained by replacing entities in aligned and conflicting scenarios with random strings (such as 'cat' with 'nsjf'). Aligned and conflicting datapoints are from the Syllogistic Dataset \cite{bertolazzi2024systematicanalysislargelanguage}.

Across all models, we see that aligned datapoints perform better than gibberish datapoints which perform better than violate datapoints, highlighting the reliance of model on its internal knowledge and inability to reason purely based on logical rules. CoT prompting consistently improves accuracy without altering the general trend observed in belief sensitivity. Interestingly, few-shot prompting does not universally help: OpenAI models (e.g., gpt-4o-mini) actually show degraded performance in few-shot settings across all belief types, suggesting potential sensitivity to in-context demonstrations or prompt formatting. In contrast, most other models maintain or slightly improve their performance under few-shot.

\subsection{Perturbing with Model-Generated Evidence}

To evaluate whether language models reason purely based on logical structure or are influenced by surface-level content, we perform an experiment involving model-generated evidence. Specifically, we prompt the models to generate evidences for a given conclusions and premises: one that supports the conclusion and one that neagtes it. The model is given complete freedom in how it constructs this evidence, encouraging creativity and variability in content. This is adapted from \cite{xie_adaptive_2024}

We then construct a separate task: given a premise, above generated evidence, and a conclusion, we ask whether the conclusion follows from the premise purely logically. The correct answer is determined solely based on the logical relation between the premise and conclusion, independent of the evidence. However, our results reveal a clear pattern: models show an increase in accuracy for logically valid datapoints when the supporting evidence aligns with the conclusion, and a drop in accuracy when the evidence contradicts it. This behavior suggests that models are not performing strict logical reasoning, but are instead heavily influenced by the factuality of premises and conclusions, even when it is explicitly stated to be potentially fabricated. This indicates a reliance on heuristic signals rather than formal logical inference.

\begin{figure*}
  \centering
  \includegraphics[width=\textwidth]{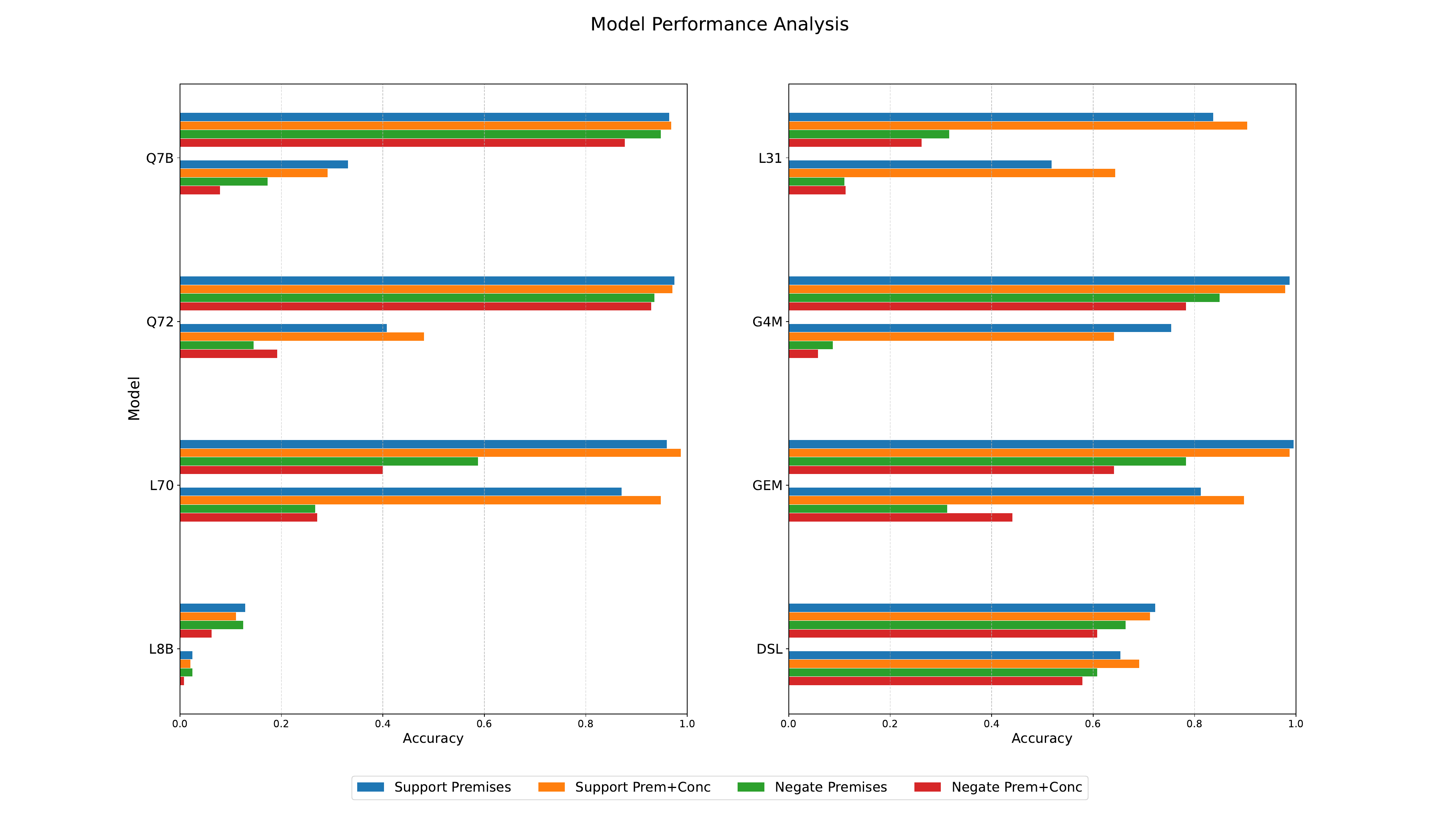}
  \caption{\textbf{Effect of evidences on logical reasoning.} Shows that providing evidences for premises or conclusions (or both) effect the outputs generated by the models even when prompted not to pay attention to factuality.}
  \label{fig:ablation-zero}
\end{figure*}

\subsection{Quantifying Counterfactual Bias of Models}

\begin{figure}[ht]
    \centering
    \includegraphics[width=1\linewidth]{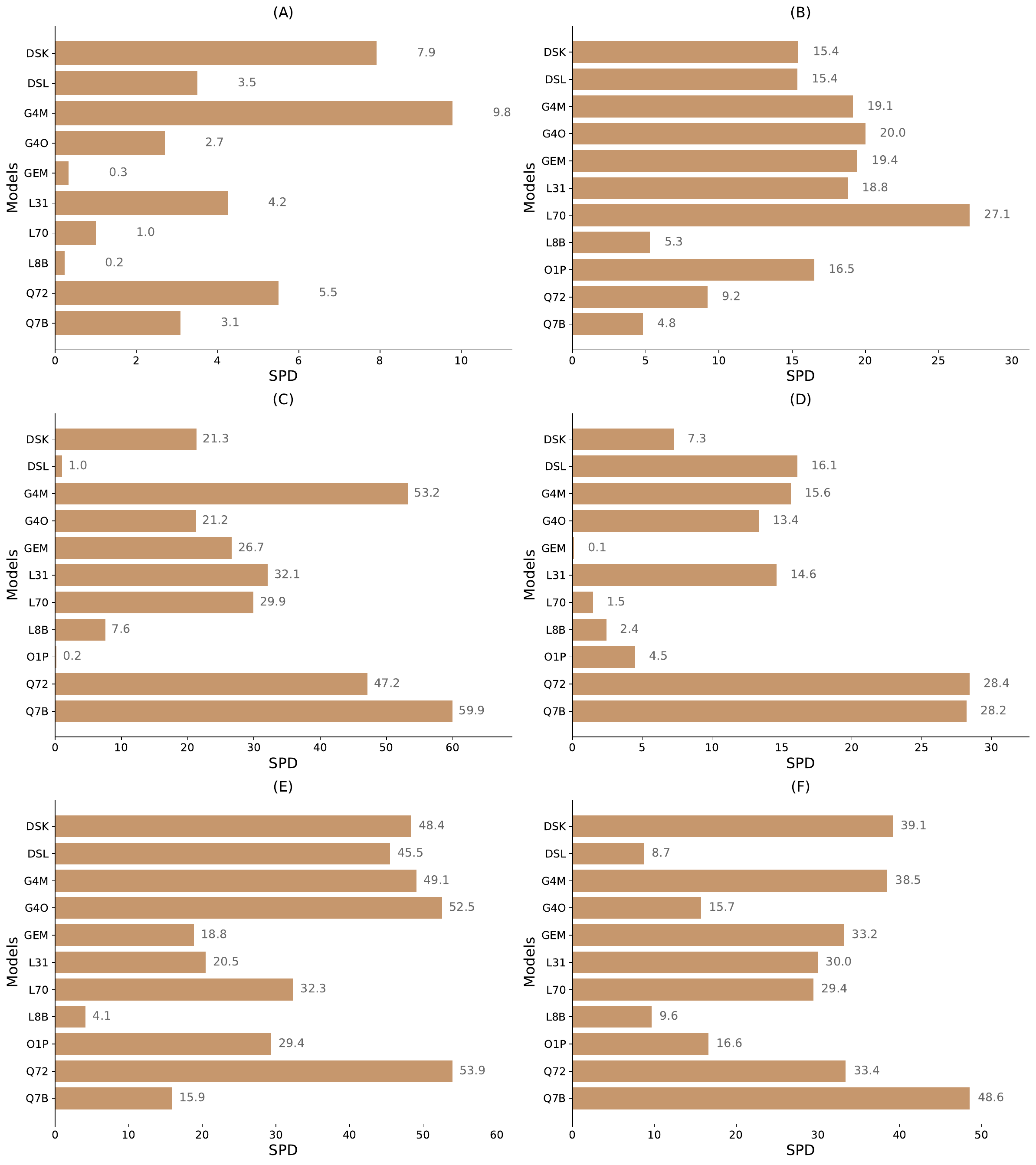}
    \captionsetup{width=0.9\linewidth}
    \caption{\textbf{SPD values for various datasets.}The Plot shows SPD values of Different models across various Datasets. We can see that the values are largely positive indicating an inherent bias towards factual answers over belief in the premises.}
    \label{fig:spd_values}
\end{figure}

We quantify the Counterfactual Bias of a model by defining Statistical Parity Difference(SPD) \cite{10.1145/3194770.3194776}. It is the difference between the likelihood that a model will give a positive answer in a scenario consistent with its factual knowledge (believable scenario) and the scenario inconsistent with its factual knowledge (unbelievable), disregarding the Logical Reasoning.

\begin{align}
\textit{SPD} ={}& (\mathbb{P}(\text{Yes} \mid \text{Believable}) \nonumber \\
               & - \mathbb{P}(\text{Yes} \mid \text{Unbelievable})) \times 100
\end{align}

A zero value for SPD implies no bias and a positive value denotes the likelihood of the model answering factually while disregarding the premises. The value of the metric is independent of the underlying ground truth distribution, which evaluates logical correctness. Hence, this metric is a good fit to evaluate all the different types of datasets.

 SPD values on various datasets are given in Figure \ref{fig:spd_values}. We can see large positive values across all the datasets indicating a large bias towards aligning with the factually correct answer even when it is not necessarily true in the given context.

\subsection{Single Prompt \textit{FaR}}
\label{app:single-far}
In figure \ref{fig:selfseg-weird-accuracy-comparison}, we evaluate results on a modified version of \textit{FaR}, where the \textit{FaR} prompt and the evaluation prompt are presented together. This setup does not exhibit any performance gains over the baseline unlike the standard \textit{FaR}, in which the \textit{FaR} prompt and the reasoning prompt are provided separately.

\section{Open Questions in Knowledge-Conflict Resolution}
\label{app:open_questions}

Our empirical findings raise fundamental questions about the mechanisms of knowledge interference in LLMs and potential resolution strategies. While our interventions show promise, they highlight critical gaps in understanding how logical reasoning interacts with parametric knowledge. Below we outline key open problems informed by our results and existing literature.

\subsection{Competing Pathways in Knowledge Interference}
\label{subsec:pathways}

The observed performance patterns suggest a potential tension between two systems that warrants further investigation:

\textbf{Hypothesis 1 (Dual-Path Interference):} Reasoning failures in counterfactual scenarios may stem from competition between:
\textit{Parametric Knowledge Activation}: Automatic retrieval of factual information acquired during pre-training (as observed in knowledge utilization studies \cite{10273594}) and
\textit{Contextual Reasoning Processes}: Controlled application of logical operations to current premises.

This hypothesized conflict aligns with findings from mechanistic analyses of LLM reasoning \cite{hong2025impliesbcircuitanalysis}, though the exact neural implementations remain unknown. Future work could employ pathway analysis techniques \cite{KUGARAJEEVAN2025125381} to test whether distinct neural substrates mediate these processes.

\subsection{Metacognitive Dissonance in Artificial Systems}
\label{subsec:dissonance}

Our believability assessment intervention parallels human cognitive dissonance resolution strategies \cite{doi:https://doi.org/10.1002/9781405165518.wbeosc058.pub2}, raising questions about whether LLMs experience analogous conflicts. Recent work demonstrates that LLMs exhibit susceptibility to influence patterns resembling human cognitive dissonance \cite{unknown}, suggesting:

\textit{Open Question:} Do LLMs develop internal conflict states when parametric knowledge contradicts task premises, and if so, how do resolution strategies differ from human cognition?

Preliminary evidence from our error analysis shows models occasionally produce self-contradictory explanations (e.g., "Although the premises state X, we know Y..."), hinting at unresolved internal conflicts. Systematic investigation using contradiction detection paradigms \cite{mundler2024selfcontradictory} could quantify this phenomenon.

\subsection{Epistemic Compartmentalization Mechanisms}
\label{subsec:compartment}

The effectiveness of metacognitive prompts suggests LLMs might transiently suppress parametric knowledge during reasoning tasks. This observation intersects with emerging research on Knowledge editing techniques that modify specific model parameters \cite{rozner2024knowledgeeditinglanguagemodels} and Hippocampal indexing approaches for contextual knowledge isolation \cite{gutiérrez2025hipporagneurobiologicallyinspiredlongterm}

\textit{Research Direction:} Can we identify neural mechanisms that enable temporary belief suspension? Layer-wise relevance propagation studies \cite{hong2025impliesbcircuitanalysis} could reveal whether metacognitive prompts induce distinct activation patterns in higher transformer layers.

\subsection{Parametric Knowledge as Malleable Belief Systems}
\label{subsec:belief}

Our evidence-based reinforcement results hint at an unexpected plasticity in LLM "beliefs." This aligns with findings that LLMs exhibit illusory truth effects \cite{unknown} and can transiently adopt fictional premises when provided authoritative citations \cite{Huang_2025}.

\textit{Open Question:} To what extent does parametric knowledge function as a belief system versus a fixed retrieval database? Controlled studies using contradiction learning paradigms \cite{zhang2024contrasolverselfalignmentlanguagemodels} could probe the conditions under which LLMs update or maintain conflicting knowledge.

\begin{figure}
\centering
\captionsetup{width=\linewidth}

\includegraphics[width=1\linewidth]{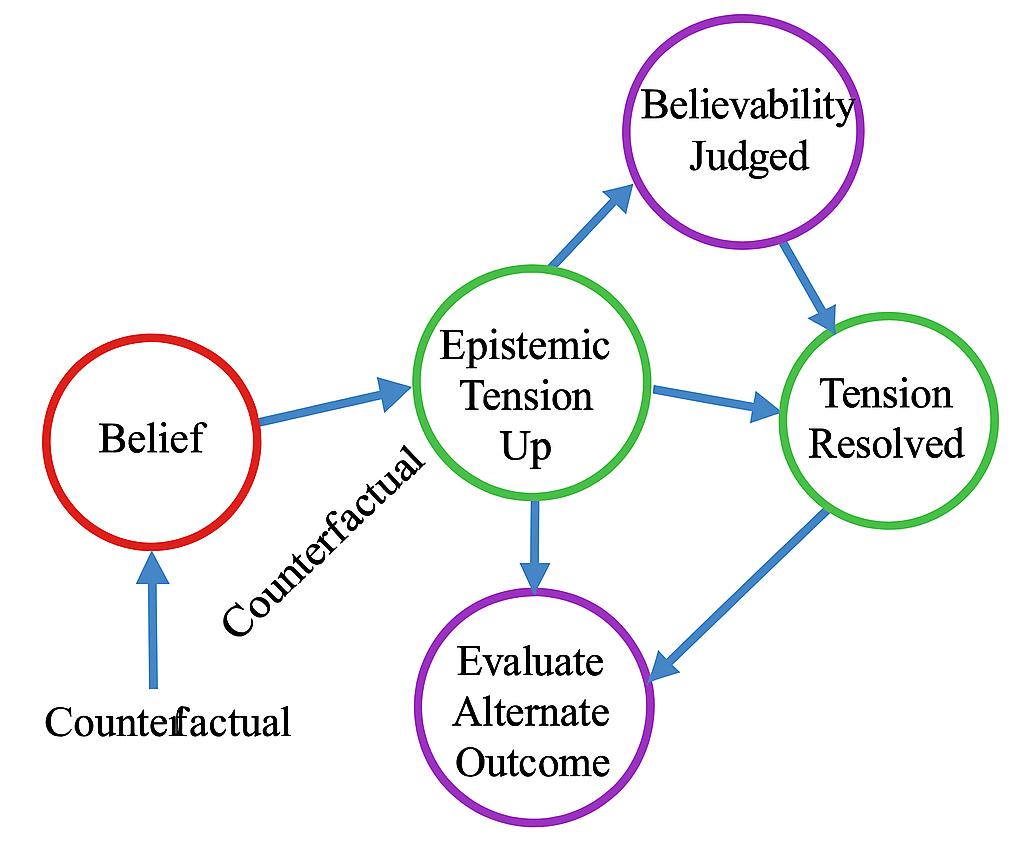}
\caption{Believability-Guided Cognitive Dissonance Resolution in Counterfactual Reasoning. Conflict between a model's stored knowledge and given premises creates dissonance. Believability assessments help resolve this dissonance by guiding the model toward epistemically appropriate inference strategies.}
\label{fig:knot-explicit}
\end{figure}

\section{Thought Anchor Analysis}
\label{app:thought-anchors}
More insights on how reasoning changes with \textit{FaR} Intervention.  The average response is 17.9 sentences long (95\% CI: [7.0, 28.8]; this corresponds to 889 tokens [95\% CI: 625, 1154]).


Based on the framework presented by ~\cite{venhoff2025understandingreasoningthinkinglanguage, bogdan2025thoughtanchorsllmreasoning} we label sentences into the sentence categories that capture reasoning functions in logical reasoning as shown in Table~\ref{tab:sentence_taxonomy}. We use Algorithm~\ref{alg:importance} to measure sentence to sentence importance.

\begin{table*}[!ht]
\centering
\small
\begin{tabular}{p{3.5cm}p{11cm}}
\toprule
\textbf{Category} & \textbf{Function} \\
\midrule
\textbf{Problem Setup} & Repeating or rephrasing the problem (initial reading or comprehension). \\[4pt]
\textbf{Plan Generation} & Stating or deciding on a plan of action (often meta-reasoning). \\[4pt]
\textbf{Fact Retrieval} & Recalling facts or problem details (without immediate computation). \\[4pt]
\textbf{Active Computation} & Performing simplifications, deductions, or derivative sentences toward the answer. \\[4pt]
\textbf{Result Consolidation} & Aggregating intermediate results, summarizing, or preparing the final answer. \\[4pt]
\textbf{Uncertainty Management} & Expressing confusion, knowledge-conflicts, re-evaluating, or proposing alternative plans. \\[4pt]
\textbf{Final Answer Emission} & Explicitly stating the final boxed answer or earlier chunks that contain the final answer. \\[4pt]
\textbf{Self Checking} & Verifying previous steps or reconfirmations. \\
\bottomrule
\end{tabular}
\caption{Sentence taxonomy with reasoning functions in problem-solving}
\label{tab:sentence_taxonomy}
\end{table*}

\begin{figure*}[!ht]
\centering
\captionsetup{width=\linewidth}

\begin{subfigure}[b]{0.48\linewidth}
    \includegraphics[width=\linewidth]{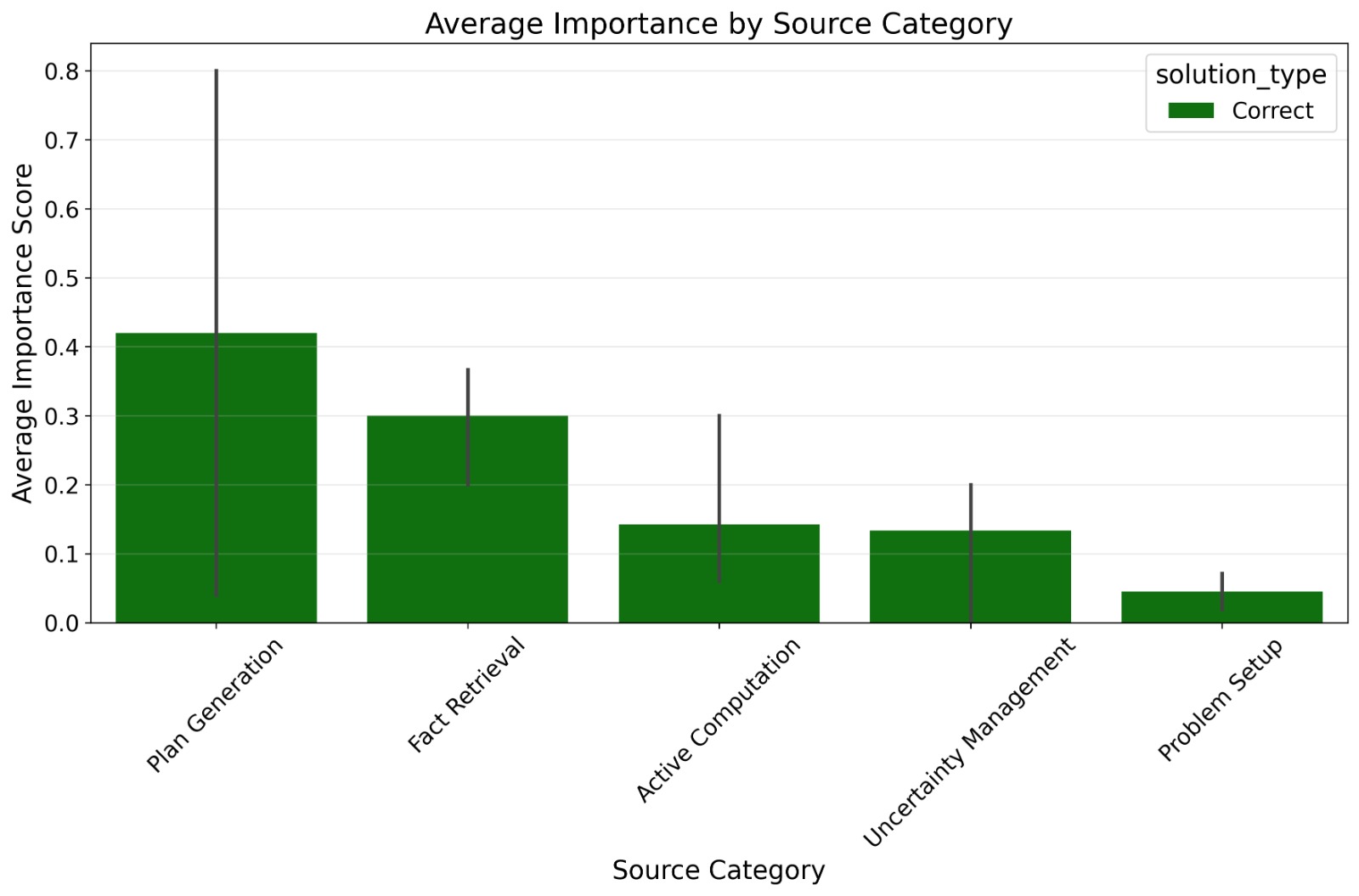}
    \caption{Normal reasoning}
    \label{fig:nonfar_avg_imp}
\end{subfigure}
\hfill
\begin{subfigure}[b]{0.48\linewidth}
    \includegraphics[width=\linewidth]{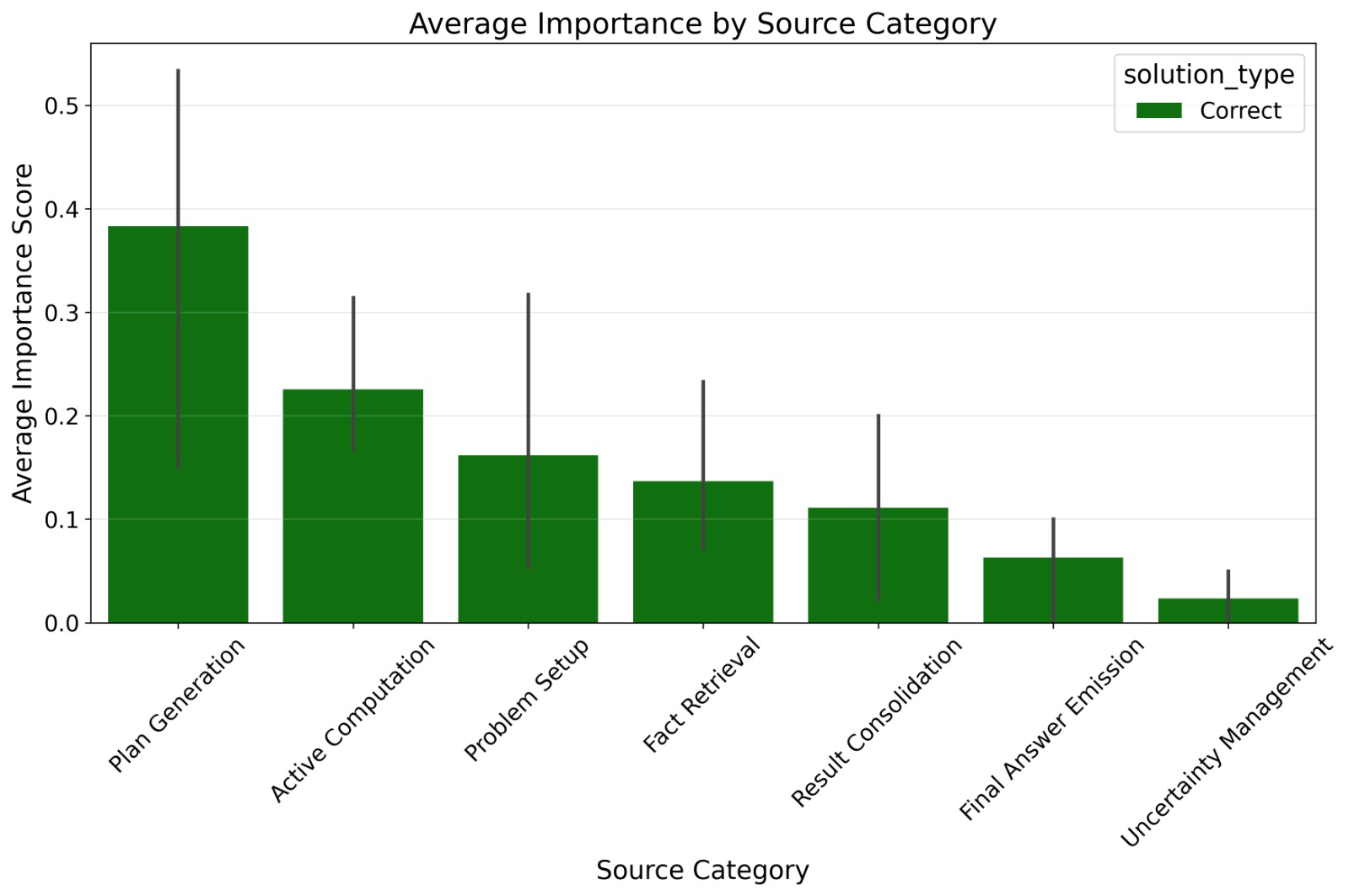}
    \caption{Reasoning with \textit{FaR}}
    \label{fig:far_avg_imp}
\end{subfigure}

\caption{ Comparison of average importance across different categories for normal reasoning vs. reasoning with \textit{FaR}.}
\label{fig:avg_imp_comparison}
\end{figure*}

We find that \textit{Plan Generation} and \textit{Fact Retrieval} sentences to be the most important sentences while reasoning normally. With \textit{Uncertainty Management} also having significant impact. While  \textit{Plan Generation}, \textit{Active Computation} and \textit{Thinking} sentences to be most influential when reasoning using the \textit{FaR} method. (Figure ~\ref{fig:nonfar_avg_imp} and ~\ref{fig:far_avg_imp}).

\begin{algorithm*}[!t]
\caption{Computation of Importance Score $\mathrm{importance}(i \rightarrow j)$}
\label{alg:importance}
\begin{algorithmic}[1]
\Require Sentence indices $i$, $j$ where $j > i$; threshold $t = 0.8$
\Ensure Importance score $\mathrm{importance}(i \rightarrow j)$

\State Obtain rollouts $R_{\mathrm{keep}}$ where sentence $i$ is kept (resampling from $i{+}1$)
\State Obtain rollouts $R_{\mathrm{remove}}$ where sentence $i$ is removed (resampling from $i$)

\For{each rollout $r \in R_{\mathrm{keep}}$}
    \State Extract all sentences $S_r$ from rollout $r$
    \State Compute sentence embeddings for all $s \in S_r$ and target sentence $j$
    \State Compute cosine similarity $\mathrm{sim}(s, j)$ for each $s \in S_r$
    \State $s^* \gets \arg\max_{s \in S_r} \mathrm{sim}(s, j)$
    \If{$\mathrm{sim}(s^*, j) \ge t$}
        \State Add $s^*$ to $\mathrm{matches}_{\mathrm{keep}}$
    \EndIf
\EndFor

\For{each rollout $r \in R_{\mathrm{remove}}$}
    \State Extract all sentences $S_r$ from rollout $r$
    \State Compute sentence embeddings and similarities $\mathrm{sim}(s, j)$ for each $s \in S_r$
    \State $s^* \gets \arg\max_{s \in S_r} \mathrm{sim}(s, j)$
    \If{$\mathrm{sim}(s^*, j) \ge t$}
        \State Add $s^*$ to $\mathrm{matches}_{\mathrm{remove}}$
    \EndIf
\EndFor

\State Compute match rates:
\State $\mathrm{match\_rate}_{\mathrm{keep}} = \dfrac{|\mathrm{matches}_{\mathrm{keep}}|}{|R_{\mathrm{keep}}|}$
\State $\mathrm{match\_rate}_{\mathrm{remove}} = \dfrac{|\mathrm{matches}_{\mathrm{remove}}|}{|R_{\mathrm{remove}}|}$

\State \Return $\mathrm{importance}(i \rightarrow j) = 
\mathrm{match\_rate}_{\mathrm{keep}} - \mathrm{match\_rate}_{\mathrm{remove}}$
\end{algorithmic}
\end{algorithm*}

\begin{figure}
\centering
\captionsetup{width=\linewidth}

\includegraphics[width=1\linewidth]{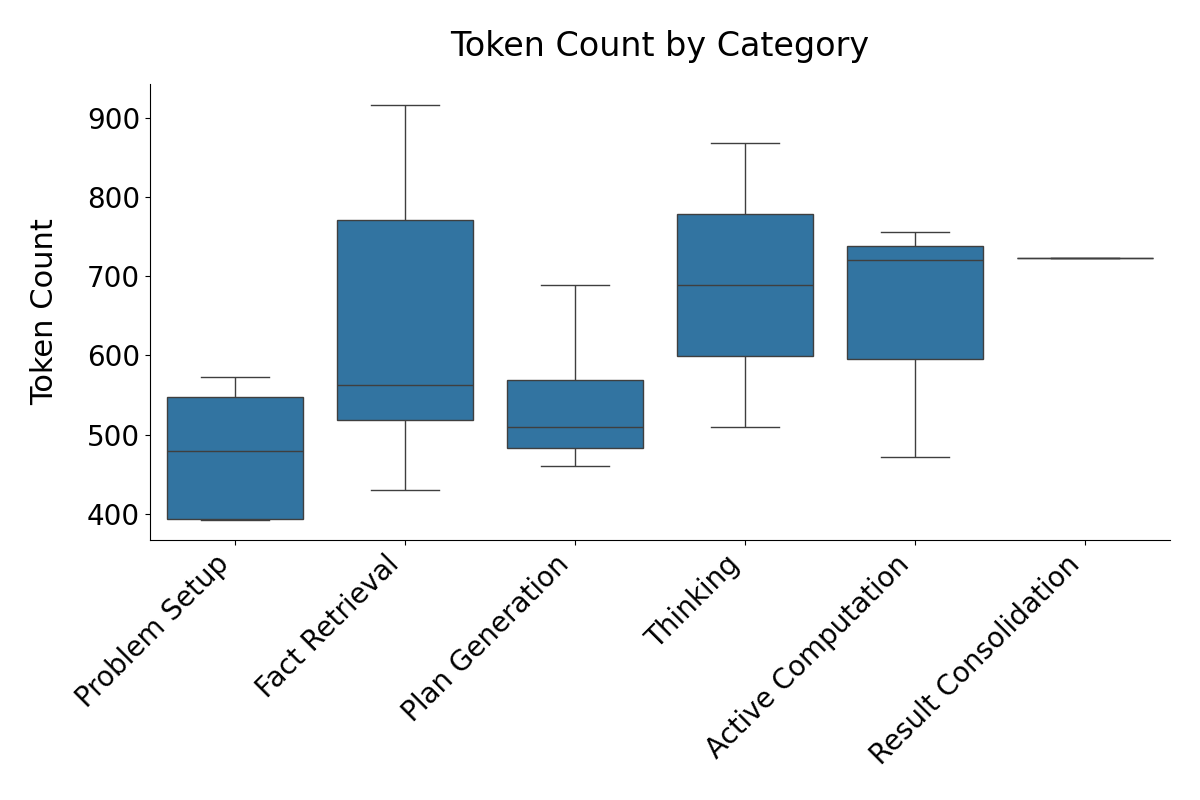}
\caption{Average Token Count by Category when reasoning normally}
\label{fig:token_n}
\end{figure}

\begin{figure}
\centering
\captionsetup{width=\linewidth}

\includegraphics[width=1\linewidth]{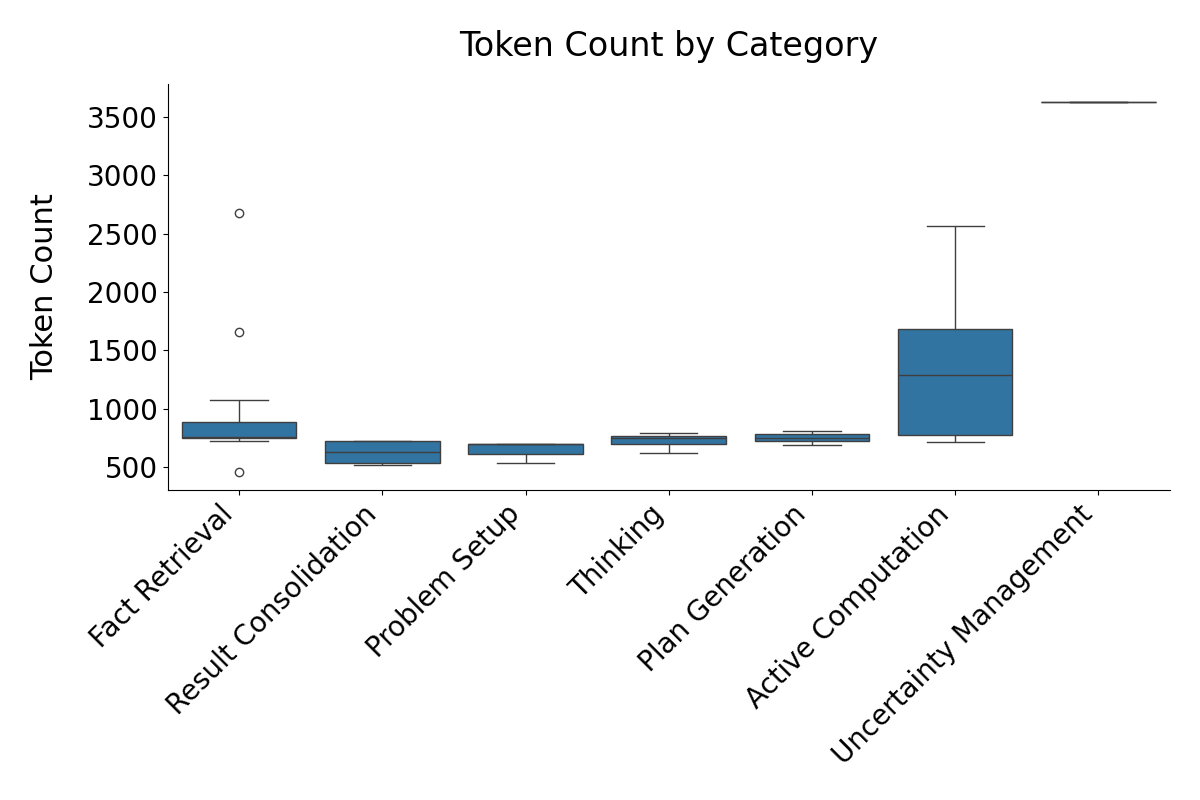}
\caption{Average Token Count by Category when reasoning with \textit{FaR}}
\label{fig:token_i}
\end{figure}

We studied examples where \textit{FaR} was helpful in turning models' incorrect answers correct. On average we see that the ration of tokens spent on \textit{Problem Setup}, \textit{Fact Retrieval} and \textit{Uncertainty Management} is much higher when reasoning normally (Figure~\ref{fig:token_n}). On the other hand, \textit{Active Computation} uses the maximum share of tokens and arrive at the correct answer as opposed to being confused with fact retrieval (Figure~\ref{fig:token_i}).

\begin{figure*}
    \centering
    \begin{subfigure}[t]{0.48\linewidth}
        \centering
        \includegraphics[width=\linewidth]{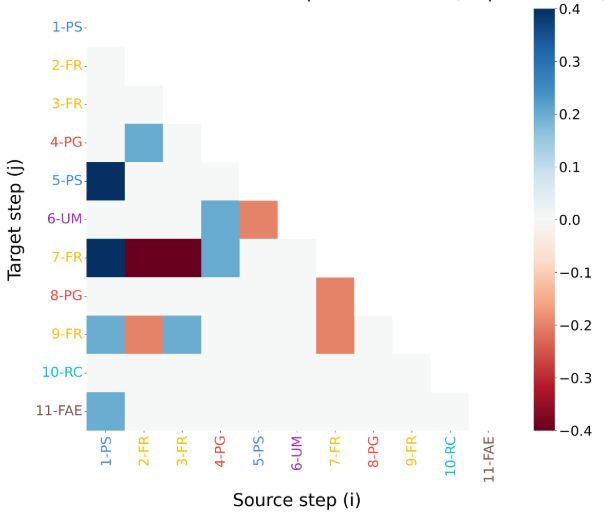}
        \caption{Steps Importance Matrix for Problem A which is \textit{Knowledge-aligned}}
        \label{fig:problemA_matrix}
    \end{subfigure}
    \hfill
    \begin{subfigure}[t]{0.48\linewidth}
        \centering
        \includegraphics[width=\linewidth]{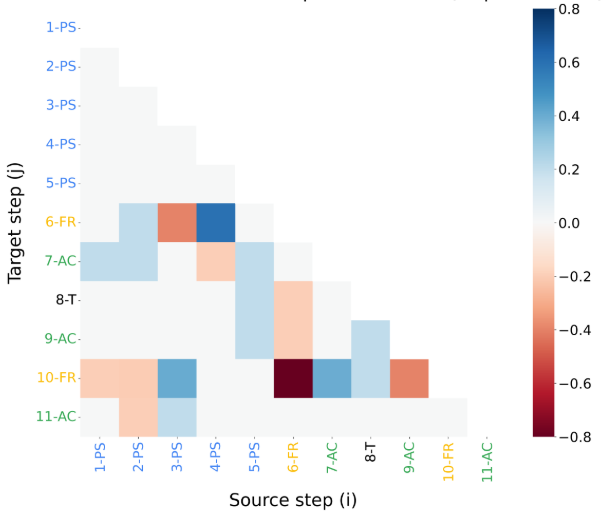}
        \caption{Steps Importance Matrix for Problem B which is \textit{Knowledge-conflicting}}
        \label{fig:problemB_matrix}
    \end{subfigure}
    \captionsetup{width=\linewidth}
    \caption{
        Comparison of step importance matrices across reasoning cases. 
        Subfigure (a) corresponds to Problem A and (b) to Problem B.
        Each cell $(i, j)$ shows the causal importance of sentence $i$ on sentence $j$, 
        calculated as the difference in the probability that sentence $j$ semantically occurs 
        (with cosine similarity $> 0.8$) when sentence $i$ is present versus when it is resampled.
    }
    \label{fig:case_study_step_importance}
\end{figure*}

\section{Case Study : How does reasoning change with knowledge-conflicts?}

Here we analyze an example where knowledge effects are visible on the outcome of logical reasoning even when the underlying logical structure is identical.

Consider these questions - \\
Question A - 
\begin{lstlisting}[basicstyle=\ttfamily\small, breaklines=true, escapeinside={(*@}{@*)}]

It's mentioned that if all (*@\textbf{cars}@*) are considered (*@\textbf{vehicles}@*), 
then all aircraft should logically be classified as planes. 
Additionally, it's stated that some aircraft are not actually planes.

Are there any (*@\textbf{cars}@*) that aren't considered (*@\textbf{vehicles}@*)?
\end{lstlisting}

Question B - 
\begin{lstlisting}[basicstyle=\ttfamily\small, breaklines=true, escapeinside={(*@}{@*)}]

It's mentioned that if all (*@\textbf{vehicles}@*) are considered (*@\textbf{cars}@*), 
then all aircraft should logically be classified as planes. 
Additionally, it's stated that some aircraft are not actually planes.

Are there any (*@\textbf{vehicles}@*) that aren't considered (*@\textbf{cars}@*)?
\end{lstlisting}

While the logical structures for these questions are identical, \textit{deepseek-r1-distill-llama-8b} returns \textit{No} for Question A while \textit{Yes} for Question B upon multiple sampling.

Since the only thing that changed is the order in which the entities \textit{cars} and \textit{vehicles} occur, the supposed difference is that answers must come from their hierarchical relationship. 

\begin{table*}[!ht]
\centering
\small
\begin{tabular}{p{0.5cm}p{10cm}p{4cm}}
\toprule
\textbf{Idx} & \textbf{Chunk Text} & \textbf{Function Tags} \\
\midrule
0 & Let's break this down: & \texttt{problem\_setup} \\[4pt]
1 & 1. The initial statement says, ``if all cars are considered vehicles, then all aircraft should logically be classified as planes.'' This is a conditional statement (if-then). & \texttt{problem\_setup} \\[4pt]
2 & It does not assert that all cars are vehicles or that all aircraft are planes; it merely states a hypothetical scenario and its logical consequence. & \texttt{fact\_retrieval} \\[4pt]
3 & 2. The statement ``some aircraft are not actually planes'' directly contradicts the hypothetical scenario where all aircraft are classified as planes. & \texttt{fact\_retrieval} \\[4pt]
4 & However, this contradiction does not negate the conditional statement itself but rather indicates that the condition under which all aircraft would be planes is not met. & \texttt{plan\_generation} \\[4pt]
5 & 3. The question asks if there are any cars that aren't considered vehicles. & \texttt{problem\_setup} \\[4pt]
6 & The information given does not directly address this question. & \texttt{uncertainty\_management} \\[4pt]
7 & It presents a hypothetical scenario about cars and vehicles and provides a separate statement about aircraft and planes. & \texttt{fact\_retrieval} \\[4pt]
8 & 4. Logically, the truth of the statements about aircraft and planes does not affect the classification of cars as vehicles. & \texttt{plan\_generation} \\[4pt]
9 & The initial conditional statement does not assert that all cars are vehicles; it simply states a hypothetical scenario. & \texttt{fact\_retrieval} \\[4pt]
10 & 5. Since the given information does not provide any evidence or logical pathway to conclude that there are cars that aren't considered vehicles, we cannot logically deduce an affirmative answer. & \texttt{result\_consolidation} \\[4pt]
11 & Therefore, based on the logical analysis of the given statements, the answer to the question ``Are there any cars that aren't considered vehicles?'' is not supported by the information provided, leading to a conclusion that we cannot affirmatively say there are cars not considered vehicles based on the given context. & \texttt{final\_answer\_emission} \\
\bottomrule
\end{tabular}
\caption{Reasoning chain for Problem A.}
\label{tab:reasoning_chain_A}
\end{table*}

\begin{table*}[!ht]
\centering
\small
\begin{tabular}{p{0.5cm}p{10cm}p{4cm}}
\toprule
\textbf{Idx} & \textbf{Chunk Text} & \textbf{Function Tags} \\
\midrule
0 & The statement is, if all vehicles are considered cars, then all aircraft should logically be classified as planes. & \texttt{problem\_setup} \\[4pt]
1 & This is a conditional statement, which can be rewritten as: if all vehicles are cars, then all aircraft are planes. & \texttt{problem\_setup} \\[4pt]
2 & Now, it's also stated that some aircraft are not actually planes. & \texttt{problem\_setup} \\[4pt]
3 & This is the second piece of information. & \texttt{problem\_setup} \\[4pt]
4 & The question is asking us if there are any vehicles that aren't considered cars. & \texttt{problem\_setup} \\[4pt]
5 & In other words, we are looking for the opposite of the first part of the conditional statement. & \texttt{problem\_setup} \\[4pt]
6 & The second piece of information is telling us that the second part of the conditional statement is false (some aircraft are not planes). & \texttt{fact\_retrieval} \\[4pt]
7 & This means we cannot conclude the first part of the conditional statement is false (all vehicles are cars). & \texttt{active\_computation} \\[4pt]
8 & In fact, this is an example of the logical fallacy of denying the antecedent. & \texttt{thinking} \\[4pt]
9 & We cannot logically conclude the opposite of the first part of the conditional statement from the opposite of the second part of the conditional statement. & \texttt{active\_computation} \\[4pt]
10 & However, we are given that if all vehicles are considered cars, then all aircraft should logically be classified as planes. & \texttt{fact\_retrieval} \\[4pt]
11 & Since we are given that some aircraft are not actually planes, then it must be the case that not all vehicles are cars. & \texttt{active\_computation} \\
\bottomrule
\end{tabular}
\caption{Reasoning chain for Problem B.}
\label{tab:reasoning_chain_B}
\end{table*}

Reasoning Chains for A and B are listed in Tables ~\ref{tab:reasoning_chain_A} and ~\ref{tab:reasoning_chain_B}. For Problem A, the top influencing steps were: Step~0 (\textit{Problem Setup}), Step~3 (\textit{Fact Retrieval}), and Step~4 (\textit{Plan Generation}). 
For Problem B, the top influencing steps were: Step~T (\textit{Active Computation}), Step~6 (\textit{Fact Retrieval}), and Step~4 (\textit{Problem Setup}). 
We compare the reasoning processes across problems based on their semantic composition and category balance. Figure ~\ref{fig:case_study_step_importance} shows the importances

Problem A shows a balanced mix of \textit{Problem Setup}, \textit{Fact Retrieval}, and \textit{Uncertainty Management}, indicating both contextual understanding and factual grounding.
In contrast, Problem~B is dominated by \textit{Active Computation} and \textit{Fact Retrieval} steps across both influential and dependent stages, suggesting a stronger focus on recalling external facts the problem correctly rather than interpreting and evaluating.

\end{document}